\documentclass[11pt,a4paper]{article}
\usepackage[hyperref]{naaclhlt2019}
\usepackage{times,latexsym}
\usepackage{url}
\usepackage[T1]{fontenc}
\usepackage{float}
\usepackage{enumitem}
\usepackage{amsmath,amsfonts,amssymb,amsthm}
\usepackage{bbm}
\usepackage{verbatim}
\usepackage{todonotes}
\usepackage{tabularx}
\usepackage{subcaption}
\usepackage{pgfplotstable}
\usepackage{booktabs}
\pgfplotsset{compat=1.7}
\usepackage{algorithm}
\usepackage{algpseudocode}
\usepackage{datatool}
\usepackage{bashful}

\usepackage{xcolor}
\usepackage{amssymb}
\usepackage{wasysym}
\usepackage{siunitx}

\aclfinalcopy 

\title{Attention is not Explanation} 

\author{Sarthak Jain \\
  Northeastern University \\
  {\small\tt jain.sar@husky.neu.edu} \\\And
  Byron C. Wallace \\
  Northeastern University \\
  {\small\tt b.wallace@northeastern.edu} \\}

\date{}

\newcommand{\vect}[1]{\boldsymbol{\mathbf{#1}}}
\newcommand{\R}{\mathbb{R}}
\newcommand{\gradient}[2]{\frac{\partial #1}{\partial #2}}

\newcommand{\trunc}[1]{\num[round-mode=places,round-precision=2]{#1}}

\newcommand{\AddSubFigureNew}[4]{
\begin{subfigure}[b]{.20\linewidth}
    \includegraphics[width=\linewidth]{graph_outputs_new/#1/#2+#3.pdf}
    \caption{#4}
\end{subfigure}
}

\newcommand{\AddSubFigureForAll}[3]{
\AddSubFigureNew{#1}{sst}{lstm+tanh}{SST (BiLSTM)}
\AddSubFigureNew{#1}{sst}{#2+tanh}{SST (#3)}
\AddSubFigureNew{#1}{diab}{lstm+tanh}{Diabetes (BiLSTM)}
\AddSubFigureNew{#1}{diab}{#2+tanh}{Diabetes (#3)}
\AddSubFigureNew{#1}{snli}{lstm+tanh}{SNLI (BiLSTM)}
\AddSubFigureNew{#1}{snli}{#2+tanh}{SNLI (#3)}
\AddSubFigureNew{#1}{cnn}{lstm+tanh}{CNN-QA (BiLSTM)}
\AddSubFigureNew{#1}{babi_1}{lstm+tanh}{BAbI 1 (BiLSTM)}
}

\newcommand{\AddSubFigureScatter}[1]{
\AddSubFigureNew{#1}{sst}{lstm+tanh}{SST (BiLSTM)}
\AddSubFigureNew{#1}{sst}{cnn_1,3,5,7_+tanh}{SST (CNN)}
\AddSubFigureNew{#1}{diab}{lstm+tanh}{Diabetes (BiLSTM)}
\AddSubFigureNew{#1}{diab}{cnn_1,3,5,7_+tanh}{Diabetes (CNN)}
\AddSubFigureNew{#1}{cnn}{lstm+tanh}{CNN-QA (BiLSTM)}
\AddSubFigureNew{#1}{babi_1}{lstm+tanh}{bAbI 1 (BiLSTM)}
\AddSubFigureNew{#1}{snli}{lstm+tanh}{SNLI (BiLSTM)}
\AddSubFigureNew{#1}{snli}{cnn_1,3,5,7_+tanh}{SNLI (CNN)}
}

\newcommand{\AddRow}[2]{
\newcount\rowIdx
\dtlforcolumn{\mean}{#2_#1}{mean}%
{%
  \advance\rowIdx by 1\relax
  \dtlgetrow{grad_#1}{\rowIdx}%
  \dtlappendentrytocurrentrow{mean_#2}{\mean}%
  \dtlrecombine
}
\newcount\rowIdx
\dtlforcolumn{\std}{#2_#1}{std}%
{%
  \advance\rowIdx by 1\relax
  \dtlgetrow{grad_#1}{\rowIdx}%
  \dtlappendentrytocurrentrow{std_#2}{\std}%
  \dtlrecombine
}
\newcount\rowIdx
\dtlforcolumn{\pvalsig}{#2_#1}{pval_sig}%
{%
  \advance\rowIdx by 1\relax
  \dtlgetrow{grad_#1}{\rowIdx}%
  \dtlappendentrytocurrentrow{pval_sig_#2}{\pvalsig}%
  \dtlrecombine
}
}

\newcommand{\DTLloadgrad}[1]{
\DTLloaddb{grad_#1}{graph_outputs_new/GradientPval/#1+lstm+tanh.csv}
\DTLloaddb{loo_#1}{graph_outputs_new/pyxc-pyc_pval/#1+lstm+tanh.csv}
\DTLloaddb{grad_avg_#1}{graph_outputs_new/GradientPval/#1+average+tanh.csv}
\DTLloaddb{loo_avg_#1}{graph_outputs_new/pyxc-pyc_pval/#1+average+tanh.csv}
\AddRow{#1}{loo}
\AddRow{#1}{grad_avg}
}

\newcommand{\generaterow}[2]{
\DTLforeach{grad_#1}{\Class=Column1,\mean=mean,\std=std,\pval=pval_sig,\meanloo=mean_loo, \stdloo=std_loo, \pvalloo=pval_sig_loo, \meanavg=mean_grad_avg, \stdavg=std_grad_avg, \pvalavg=pval_sig_grad_avg}%
{%
  \DTLiffirstrow{#2}{} & \Class & \trunc{\mean} $\pm$ \trunc{\std} & \trunc{\pval} & \trunc{\meanavg} $\pm$ \trunc{\stdavg} & \trunc{\pvalavg} & \trunc{\meanloo} $\pm$ \trunc{\stdloo} & \trunc{\pvalloo} \DTLiflastrow{\hspace{-4pt}}{\\} 
}%
}

\newcommand{\GradTabs}{
\DTLloadgrad{sst}
\DTLloadgrad{imdb}
\DTLloadgrad{diab}
\DTLloadgrad{tweet}
\DTLloadgrad{agnews}
\DTLloadgrad{anemia}
\DTLloadgrad{20News_sports}
\DTLloadgrad{cnn}
\DTLloadgrad{snli}
\DTLloadgrad{babi_1}
\DTLloadgrad{babi_2}
\DTLloadgrad{babi_3}

\begin{table*}
\small
\centering
\begin{tabular}{cccccccc}
& & \multicolumn{2}{c}{Gradient (BiLSTM) $\tau_g$} & \multicolumn{2}{c}{Gradient (Average) $\tau_{g}$} & \multicolumn{2}{c}{Leave-One-Out (BiLSTM) $\tau_{loo}$}  \\ \hline
Dataset & Class & Mean $\pm$ Std. & Sig. Frac. & Mean $\pm$ Std. & Sig. Frac. & Mean $\pm$ Std. & Sig. Frac.\\ \hline
\generaterow{sst}{SST} \\
\generaterow{imdb}{IMDB} \\
\generaterow{tweet}{ADR Tweets} \\
 \generaterow{20News_sports}{20News} \\
 \generaterow{agnews}{AG News} \\
 \generaterow{diab}{Diabetes} \\
 \generaterow{anemia}{Anemia} \\ \hline
 \generaterow{cnn}{CNN} \\
 \generaterow{babi_1}{bAbI 1} \\
 \generaterow{babi_2}{bAbI 2} \\
 \generaterow{babi_3}{bAbI 3} \\ \hline
 \generaterow{snli}{SNLI} \\

\end{tabular} \vspace{-.75em}
\caption{Mean and std. dev. of correlations between gradient/leave-one-out importance measures and attention weights. \emph{Sig. Frac.} columns report the fraction of instances for which this correlation is statistically significant; note that this largely depends on input length, as correlation does tend to exist, just weakly. Encoders are denoted parenthetically. These are representative results; exhaustive results for all encoders are available to browse online.} \vspace{-1em}
    \label{tab:tau-sig}
\end{table*}
}

\begin{document}
\maketitle
\begin{abstract}
Attention mechanisms have seen wide adoption in neural NLP models. 
In addition to improving predictive performance, these are often touted as affording transparency: models equipped with attention provide a distribution over attended-to input units, and this is often presented (at least implicitly) as communicating the relative importance of inputs. 
However, it is unclear what relationship exists between attention weights and model outputs. 
In this work we perform extensive experiments across a variety of NLP tasks that aim to assess the degree to which attention weights provide meaningful ``explanations" for predictions. 
We find that they largely do not.
For example, learned attention weights are frequently uncorrelated with gradient-based measures of feature importance, and one can identify very different attention distributions that nonetheless yield equivalent predictions. 
Our findings show that standard attention modules do not provide meaningful explanations and should not be treated as though they do. Code to reproduce all experiments is available at {\small \url{https://github.com/successar/AttentionExplanation}}.
\end{abstract}

\section{Introduction and Motivation}

\emph{Attention mechanisms}  \cite{bahdanau2014neural} induce conditional distributions over input units to compose a weighted context vector for downstream modules. 
These are now a near-ubiquitous component of neural NLP architectures. 
Attention weights are often claimed (implicitly or explicitly) to afford insights into the ``inner-workings'' of models: for a given output one can inspect the inputs to which the model assigned large attention weights. 
Li \emph{et al.} \shortcite{li2016understanding} summarized this commonly held view in NLP: ``Attention provides an important way to explain the workings of neural models". Indeed, claims that attention provides interpretability are common in the literature, e.g., \cite{xu2015show,choi2016retain,lei2017interpretable,martins2016softmax,xie2017interpretable,mullenbach2018explainable}.\footnote{We do not intend to single out any particular work; indeed one of the authors has himself presented (supervised) attention as providing interpretability \cite{zhang2016rationale}.}

\begin{figure}
    \includegraphics[scale=0.425]{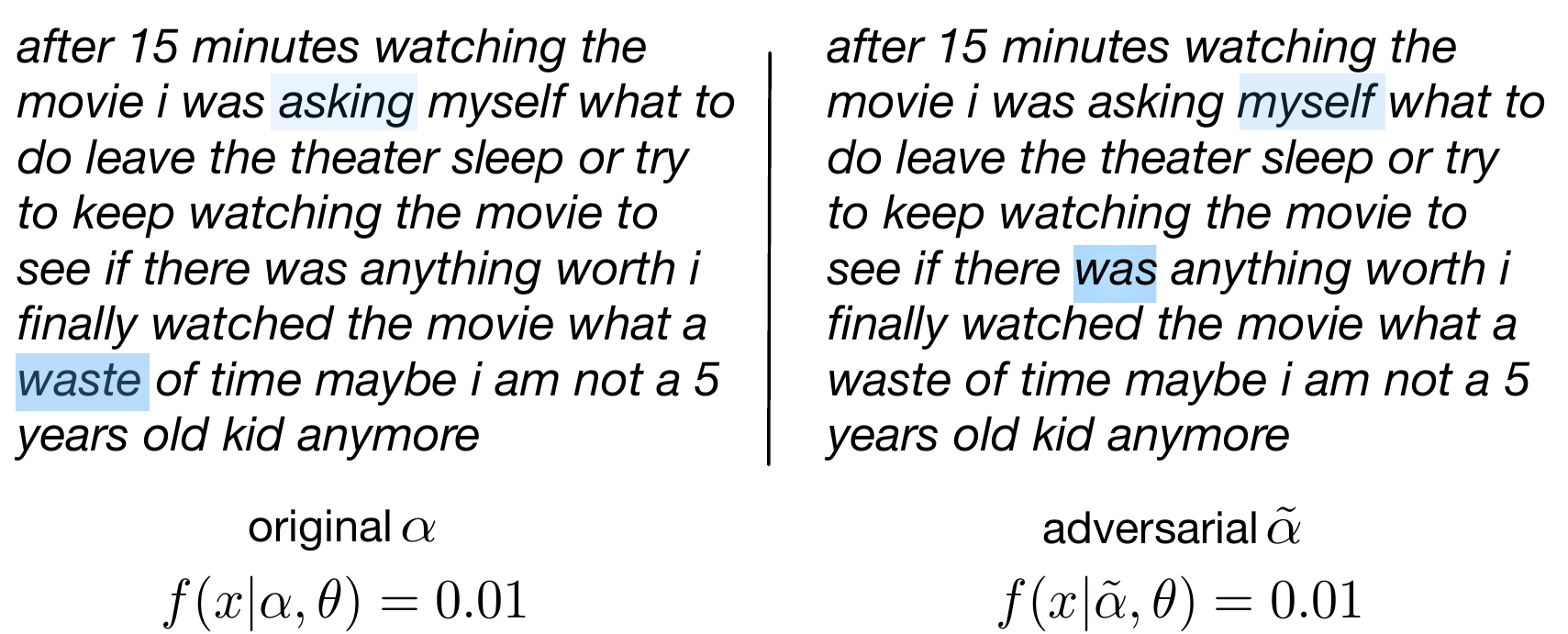}
    \caption{Heatmap of attention weights induced over a negative movie review. We show observed model attention (left) and an adversarially constructed set of attention weights (right). Despite being quite dissimilar, these both yield effectively the same prediction (0.01).} 
    \label{fig:ad-attn}
\end{figure}

Implicit in this is the assumption that the inputs (e.g., words) accorded high attention weights are responsible for model outputs. 
But as far as we are aware, this assumption has not been formally evaluated. 
Here we empirically investigate the relationship between attention weights, inputs, and outputs. 

Assuming attention provides a faithful explanation for model predictions, we might expect the following properties to hold. 
(i) Attention weights should correlate with feature importance measures (e.g., gradient-based measures);
(ii) Alternative (or \emph{counterfactual}) attention weight configurations ought to yield corresponding changes in prediction (and if they do not then are equally plausible as explanations).
We report that neither property is consistently observed by a BiLSTM with a standard attention mechanism in the context of text classification, question answering (QA), and Natural Language Inference (NLI) tasks.

Consider Figure \ref{fig:ad-attn}.
The left panel shows the original attention distribution $\alpha$ over the words of a particular movie review using a standard attentive BiLSTM architecture for sentiment analysis.
It is tempting to conclude from this that the token \emph{waste} is largely responsible for the model coming to its disposition of `negative' ($\hat{y} = 0.01$). 
But one can construct an alternative attention distribution $\tilde{\alpha}$ (right panel) that attends to entirely different tokens yet yields an essentially identical prediction (holding all other parameters of $f$, $\theta$, constant). 

Such counterfactual distributions imply that explaining the original prediction by highlighting attended-to tokens is misleading. 
One may, e.g., now conclude from the right panel that model output was due primarily to \emph{was}; but both \emph{waste} and \emph{was} cannot simultaneously be responsible.
Further, the attention weights in this case correlate only weakly with gradient-based measures of feature importance ($\tau_g=0.29$). 
And arbitrarily permuting the entries in $\alpha$ yields a median output difference of 0.006 with the original prediction. 

These and similar findings call into question the view that attention provides meaningful insight into model predictions, at least for RNN-based models. 
We thus caution against using attention weights to highlight input tokens ``responsible for'' model outputs and constructing just-so stories on this basis.

\vspace{.5em}
\noindent {\bf Research questions and contributions}. We examine the extent to which the (often implicit) narrative that attention provides model \emph{transparency}  \cite{lipton2016mythos}. We are specifically interested in whether attention weights indicate \emph{why} a model made the prediction that it did. This is sometimes called \emph{faithful} explanation \cite{ross2017right}. 
We investigate whether this holds across tasks by exploring the following empirical questions. 

\begin{enumerate}
\item To what extent do induced attention weights correlate with measures of feature importance -- specifically, those resulting from gradients and \emph{leave-one-out} (LOO) methods?

\item Would alternative attention weights (and hence distinct heatmaps/``explanations'') necessarily yield different predictions?
\end{enumerate} 

Our findings for attention weights in recurrent (BiLSTM) encoders with respect to these questions are summarized as follows: (1) Only weakly and inconsistently, and, (2) No; it is very often possible to construct \emph{adversarial} attention distributions that yield effectively equivalent predictions as when using the originally induced attention weights, despite attending to entirely different input features. Even more strikingly, randomly permuting attention weights often induces only minimal changes in output. By contrast, attention weights in simple, feedforward (weighted average) encoders enjoy better behaviors with respect to these criteria.

\section{Preliminaries and Assumptions}
\label{section:prelims}

We consider exemplar NLP tasks for which attention mechanisms are commonly used: classification, natural language inference (NLI), and question answering.\footnote{While attention is perhaps \emph{most} common in seq2seq tasks like translation, our impression is that interpretability is not typically emphasized for such tasks, in general.} We adopt the following general modeling assumptions and notation. 

We assume model inputs $\vect{x} \in \R^{T \times |V|}$, composed of one-hot encoded words at each position. These are passed through an embedding matrix $\vect{E}$ which provides dense ($d$ dimensional) token representations $\vect{x}_e \in \R^{T \times d}$. Next, an encoder {\bf Enc} consumes the embedded tokens in order, producing $T$ $m$-dimensional hidden states: $\vect{h} = \textbf{Enc}(\vect{x}_e) \in \R^{T \times m}$. We predominantly consider a Bi-RNN as the encoder module, and for contrast we analyze unordered `average' embedding variants in which $h_t$ is the embedding of token $t$ after being passed through a linear projection layer and ReLU activation. For completeness we also considered ConvNets, which are somewhere between these two models; we report results for these in the supplemental materials. 

\newcolumntype{Y}{>{\centering\arraybackslash}X}
\begin{table*}[!htbp]
   \small
    \begin{tabular}{l l l || c c || c}
         \emph{Dataset} & $\boldsymbol{|V|}$ & \emph{Avg. length} &
         \emph{Train size} & \emph{Test size} & \emph{Test performance (LSTM)} \\
          \hline
         SST & 16175 &  19 & 3034 / 3321 & 863 / 862 & 0.81 \\
        IMDB & 13916 & 179 & 12500 / 12500 & 2184 / 2172 & 0.88 \\
        ADR Tweets & 8686 & 20 & 14446 / 1939 & 3636 / 487 & 0.61 \\
        20 Newsgroups & 8853 & 115 & 716 / 710 & 151 / 183 & 0.94 \\
        AG News & 14752 & 36 & 30000 / 30000 & 1900 / 1900 & 0.96 \\
        Diabetes (MIMIC) & 22316 & 1858 & 6381 / 1353 & 1295 / 319 & 0.79 \\ 
        Anemia (MIMIC) & 19743 & 2188 & 1847 / 3251 & 460 / 802 & 0.92 \\ 
        \hline 
        CNN & 74790 & 761 & 380298 & 3198 & 0.64 \\ 
        bAbI (Task 1 / 2 / 3) & 40 & 8 / 67 / 421 & 10000 & 1000 & 1.0 / 0.48 / 0.62 \\ 
        \hline
        SNLI & 20982 & 14 & 182764 / 183187 / 183416 & 3219 / 3237 / 3368 & 0.78
         
    \end{tabular}
    \caption{Dataset characteristics. For train and test size, we list the cardinality for each class, where applicable: $0$/$1$ for binary classification (top), and 0 / 1 / 2 for NLI (bottom). Average length is in tokens. Test metrics are F1 score, accuracy, and micro-F1 for classification, QA, and NLI, respectively; all correspond to performance using a BiLSTM encoder. We note that results using convolutional and average (i.e., non-recurrent) encoders are comparable for classification though markedly worse for QA tasks.} 
    \label{tab:datastats}
    \vspace{-1em}
\end{table*}

A similarity function $\phi$ maps $\vect{h}$ and a query $\vect{Q}\in \R^{m}$ (e.g., hidden representation of a question in QA, or the hypothesis in NLI) to scalar scores, and attention is then induced over these: $\hat{\vect{\alpha}} = \text{softmax}(\phi (\vect{h}, \vect{Q})) \in \R^{T}$. In this work we consider two common similarity functions: \emph{Additive} $\phi(\vect{h}, \vect{Q}) = \vect{v}^T\text{tanh}(\vect{W_1} \vect{h} + \vect{W_2} \vect{Q})$ \cite{bahdanau2014neural} and \emph{Scaled Dot-Product} $\phi(\vect{h}, \vect{Q}) = \frac{\vect{h}\vect{Q}}{\sqrt{m}}$ \cite{vaswani2017attention}, where $\vect{v}, \vect{W_1}, \vect{W_2}$ are model parameters.

Finally, a dense layer \textbf{Dec} with parameters $\vect{\theta}$ consumes a weighted instance representation and yields a prediction $\hat{y} = \sigma(\vect{\theta} \cdot h_\alpha) \in \R^{|\mathcal{Y}|}$, where $h_\alpha = \sum_{t=1}^T \hat{\alpha}_t \cdot h_t$; $\sigma$ is an output activation function; and $|\mathcal{Y}|$ denotes the label set size.

\section{Datasets and Tasks}

For \textbf{binary text classification}, we use:

    \emph{Stanford Sentiment Treebank (SST)} \cite{socher2013recursive}. 10,662 sentences tagged with sentiment on a scale from 1 (most negative) to 5 (most positive). We filter out neutral instances and dichotomize the remaining sentences into positive (4, 5) and negative (1, 2).
    
    \emph{IMDB Large Movie Reviews Corpus} \cite{maas-EtAl:2011:ACL-HLT2011}. Binary sentiment classification dataset containing 50,000 polarized (positive or negative) movie reviews, split into half for training and testing.
    
    \emph{Twitter Adverse Drug Reaction} dataset \cite{doi:10.1093/jamia/ocu041}. A corpus of $\sim$8000 tweets retrieved from Twitter, annotated by domain experts as mentioning adverse drug reactions. We use a superset of this dataset.
    
    \emph{20 Newsgroups (Hockey vs Baseball)}. Collection of $\sim$20,000 newsgroup correspondences, partitioned (nearly) evenly across 20 categories. We extract instances belonging to \emph{baseball} and \emph{hockey}, which we designate as 0 and 1, respectively, to derive a binary classification task.
    
    \emph{AG News Corpus (Business vs World)}.\footnote{\url{http://www.di.unipi.it/~gulli/AG_corpus_of_news_articles.html}} 496,835 news articles from 2000+ sources. We follow \cite{zhang2015character} in filtering out all but the top 4 categories. We consider the binary classification task of discriminating between \emph{world} (0) and \emph{business} (1) articles.
    
    \emph{MIMIC ICD9 (Diabetes)} \cite{johnson2016mimic}. A subset of discharge summaries from the MIMIC III dataset of electronic health records. The task is to recognize if a given summary has been labeled with the ICD9 code for diabetes (or not).
    
    \emph{MIMIC ICD9 (Chronic vs Acute Anemia)} \cite{johnson2016mimic}. A subset of discharge summaries from MIMIC III dataset \cite{johnson2016mimic} known to correspond to patients with anemia. Here the task to distinguish the type of anemia for each report -- \emph{acute} (0) or \emph{chronic} (1).

\vspace{.75em}
\noindent For {\bf Question Answering (QA)}:

    \emph{CNN News Articles} \cite{hermann2015teaching}. A corpus of cloze-style questions created via automatic parsing of news articles from CNN. Each instance comprises a paragraph-question-answer triplet, where the answer is one of the anonymized entities in the paragraph. 
    
    \emph{bAbI} \cite{weston2015towards}. We consider the three tasks presented in the original bAbI dataset paper, training separate models for each. These entail finding (i) a single supporting fact for a question and (ii) two or (iii) three supporting statements, chained together to compose a coherent line of reasoning.

\vspace{.75em}
\noindent Finally, for {\bf Natural Language Inference} (NLI):

The \emph{SNLI dataset} \cite{bowman2015large}. 570k human-written English sentence pairs manually labeled for balanced classification with the labels \emph{neutral}, \emph{contradiction}, and \emph{entailment}, supporting the task of natural language inference (NLI). In this work, we generate an attention distribution over premise words conditioned on the hidden representation induced for the hypothesis.

We restrict ourselves to comparatively simple instantiations of attention mechanisms, as described in the preceding section. This means we do not consider recently proposed `BiAttentive' architectures that attend to tokens in the respective inputs, conditioned on the other inputs \cite{parikh2016decomposable,seo2016bidirectional,xiong2016dynamic}.

Table \ref{tab:datastats} provides summary statistics for all datasets, as well as the observed test performances for additional context. 

\GradTabs

\section{Experiments}
\label{section:experiments}

We run a battery of experiments that aim to examine empirical properties of learned attention weights and to interrogate their interpretability and transparency. The key questions are: \emph{Do learned attention weights agree with alternative, natural measures of feature importance}? And, \emph{Had we attended to different features, would the prediction have been different?}

More specifically, in Section \ref{section:gradient-correlations}, we empirically analyze the correlation between gradient-based feature importance and learned attention weights, and between emph{leave-one-out} (LOO; or `feature erasure') measures and the same. 
In Section \ref{section:counterfactuals} we then consider counterfactual (to those observed) attention distributions. 
Under the assumption that attention weights are explanatory, such counterfactual distributions may be viewed as alternative potential explanations; if these do not correspondingly change model output, then it is hard to argue that the original attention weights provide meaningful explanation in the first place.

To generate counterfactual attention distributions, we first consider randomly permuting observed attention weights and recording associated changes in model outputs (\ref{subsection:permutations}). 
We then propose explicitly searching for ``adversarial'' attention weights that maximally differ from the observed attention weights (which one might show in a heatmap and use to explain a model prediction), and yet yield an effectively equivalent prediction (\ref{subsection:adversarial}). 
The latter strategy also provides a useful potential metric for the reliability of attention weights as explanations: we can report a measure quantifying how different attention weights can be for a given instance without changing the model output by more than some threshold $\epsilon$. 

All results presented below are generated on test sets. We present results for \emph{Additive} attention below. The results for \emph{Scaled Dot Product} in its place are generally comparable. We provide a web interface to interactively browse the (very large set of) plots for all datasets, model variants, and experiment types: {\small\url{https://successar.github.io/AttentionExplanation/docs/}}.

In the following sections, we use Total Variation Distance (TVD) as the measure of change between output distributions, defined as follows. $\text{TVD}(\hat{y}_1, \hat{y}_2) = \frac{1}{2}\sum_{i=1}^{|\mathcal{Y}|} |\hat{y}_{1i}-\hat{y}_{2i}|$. We use the Jensen-Shannon Divergence (JSD) to quantify the difference between two attention distributions: $\text{JSD}(\alpha_1, \alpha_2) = \frac{1}{2} \text{KL}[\alpha_1||\frac{\alpha_1 + \alpha_2}{2}] + \frac{1}{2}\text{KL}[\alpha_2||\frac{\alpha_1 + \alpha_2}{2}]$.

\subsection{Correlation Between Attention and Feature Importance Measures}
\label{section:gradient-correlations}

\begin{figure*}
    \centering
    \AddSubFigureForAll{GradientXHist}{average}{Average}
    \vspace{-.5em}
    \caption{Histogram of \textbf{Kendall $\tau$} between attention and gradients. Encoder variants are denoted parenthetically; colors indicate predicted classes. Exhaustive results are available for perusal online.}
    \label{fig:GradientX}
\end{figure*}

\begin{figure}

    \centering
    \includegraphics[width=\columnwidth]{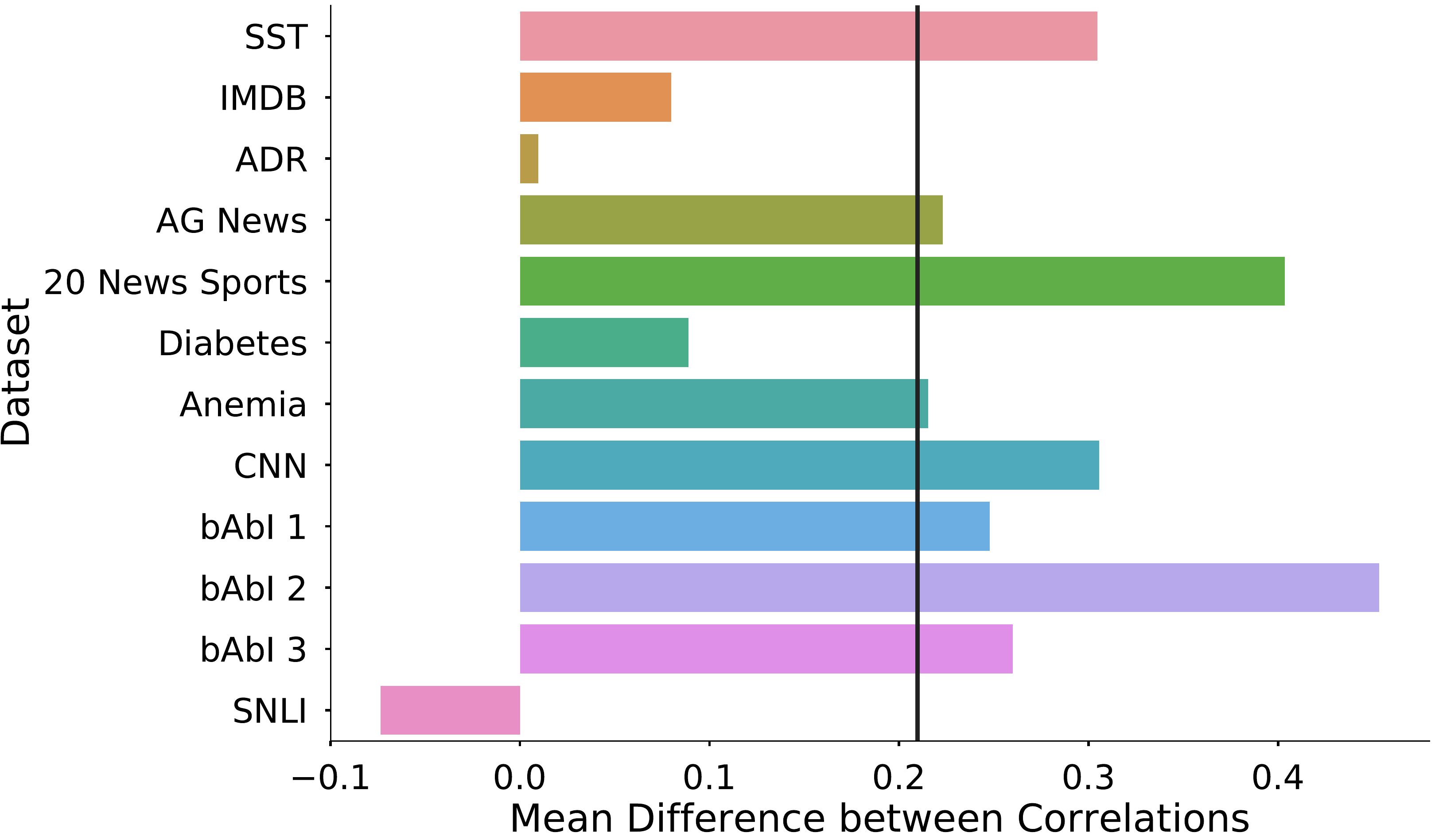}
    \caption{Mean difference in correlation of (i) LOO vs. Gradients and (ii) Attention vs. LOO scores using BiLSTM Encoder + Tanh Attention. On average the former is more correlated than the latter by $> $0.2 $\tau_{\textit{loo}}$.} 
    \label{fig:corrgl_loo}
\end{figure} 

\begin{figure}
    \centering
    \includegraphics[width=\columnwidth]{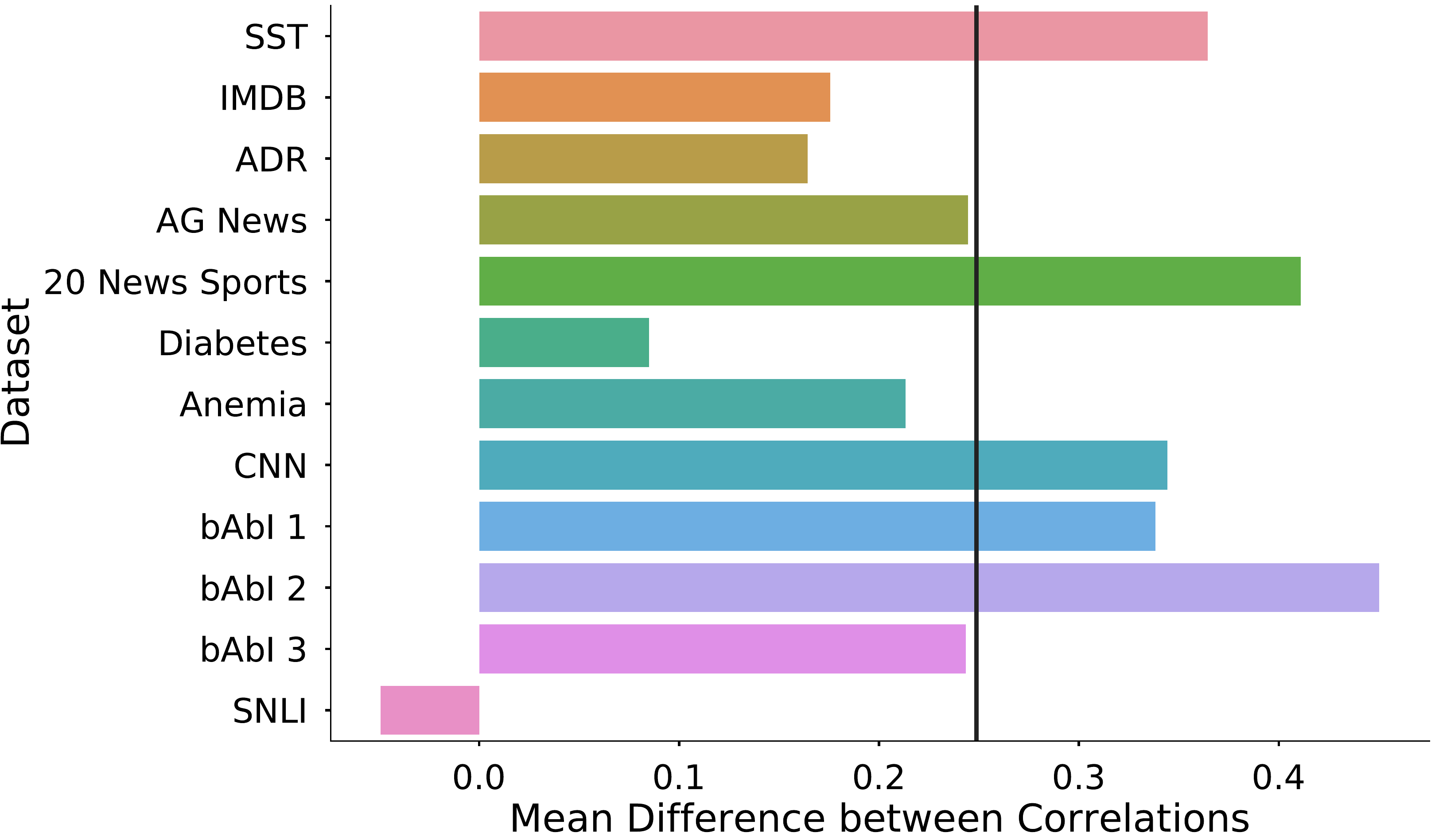}
    \caption{Mean difference in correlation of (i) LOO vs. Gradients and (ii) Attention vs. Gradients using BiLSTM Encoder + Tanh Attention. On average the former is more correlated than the latter by $\sim$0.25 $\tau_g$.}
    \label{fig:corrgl_ag}
\end{figure}

\begin{figure}
    \centering
    \includegraphics[width=\columnwidth]{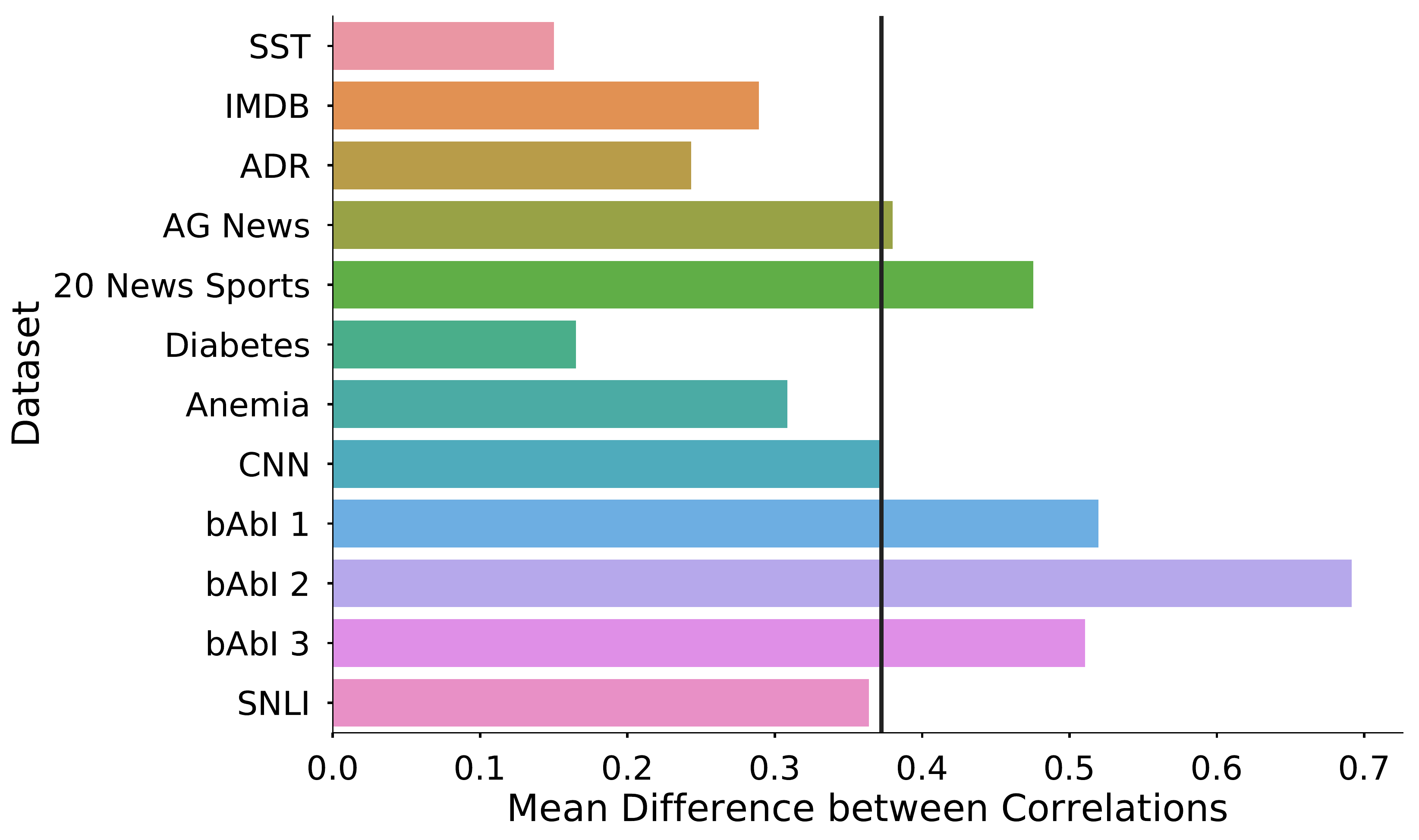}
    \caption{Difference in mean correlation of attention weights vs. LOO importance measures for (i) Average (feed-forward projection) and (ii) BiLSTM Encoders with Tanh attention. Average correlation (vertical bar) is on average $\sim$0.375 points higher for the simple feedforward encoder, indicating greater correspondence with the LOO measure.}
    \label{fig:corral}
\end{figure}

We empirically characterize the relationship between attention weights and corresponding feature importance scores. Specifically we measure correlations between attention and: (1) gradient based measures of feature importance ($\tau_g$), and, (2) differences in model output induced by leaving features out ($\tau_{\textit{loo}}$).
While these measures are themselves insufficient for interpretation of neural model behavior \cite{feng2018pathologies}, they do provide measures of individual feature importance with known semantics \cite{ross2017right}. It is thus instructive to ask whether these measures correlate with attention weights.

The process we follow to quantify this is described by Algorithm \ref{alg:gradient}. 
We denote the input resulting from removing the word at position $t$ in $\vect{x}$ by $\vect{x}_{-t}$.
Note that we disconnect the computation graph at the attention module so that the gradient does not flow through this layer: This means the gradients tell us how the prediction changes as a function of inputs, \emph{keeping the attention distribution fixed}.\footnote{For further discussion concerning our motivation here, see the Appendix. We also note that LOO results are comparable, and do not have this complication.}

\begin{algorithm}
\begin{algorithmic}
\State $\vect{h} \gets \textrm{Enc}(\vect{x})$, $\hat{\alpha} \gets \textrm{softmax}(\phi(\vect{h}, \vect{Q}))$
\State $\hat{y} \gets \textrm{Dec}(\vect{h}, \alpha)$
\State $g_t \gets |\sum_{w=1}^{|V|} \mathbbm{1}[\vect{x}_{tw} = 1]\gradient{y}{\vect{x}_{tw}}|\;,\forall t \in [1, T]$
\State $\tau_g \gets \text{Kendall-}\tau(\alpha, g)$
\State $\Delta\hat{y}_t \gets \text{TVD}(\hat{y}(\vect{x}_{-t}), \hat{y}(\vect{x}))\;,\forall t \in [1, T]$
\State $\tau_{\textit{loo}} \gets \text{Kendall-}\tau(\alpha, \Delta\hat{y})$
\end{algorithmic}
\caption{Feature Importance Computations}
\label{alg:gradient}
\end{algorithm}

Table \ref{tab:tau-sig} reports summary statistics of Kendall $\tau$ correlations for each dataset. Full distributions are shown in \autoref{fig:GradientX}, which plots histograms of $\tau_g$ for every data point in the respective corpora. 
(Corresponding plots for $\tau_{loo}$ are similar and the full set can be browsed via the aforementioned online supplement.)
We plot these separately for each class: orange (\textcolor[HTML]{d95f02}{$\blacksquare$}) represents instances predicted as positive, and purple (\textcolor[HTML]{7570b3}{$\blacksquare$}) those predicted to be negative. 
For SNLI, \textcolor[HTML]{7570b3}{$\blacksquare$}, \textcolor[HTML]{d95f02}{$\blacksquare$}, \textcolor[HTML]{1b9e77}{$\blacksquare$}, code for neutral, contradiction and entailment, respectively.

In general, observed correlations are modest (recall that a value of 0 indicates no correspondence, while 1 implies perfect concordance) for the BiLSTM model. The centrality of observed densities hovers around or below 0.5 in most of the corpora considered. 
On some datasets --- notably the MIMIC tasks, and to a lesser extent the QA corpora --- correlations are consistently significant, but they remain relatively weak. 
The more consistently significant correlations observed on these datasets is likely attributable to the increased length of the documents that they comprise, which in turn provide sufficiently large sample sizes to establish significant (if weak) correlation between attention weights and feature importance scores.

Gradients for inputs in the ``average'' (linear projection) embedding based models show much higher correspondence with attention weights, as we would expect for these simple encoders. We hypothesize that the trouble is in the BiRNN inducing attention on the basis of hidden states, which are influenced (potentially) by all words, and then presenting the scores in reference to the original input.


A question one might have here is how well correlated LOO and gradients are with \emph{one another}. We report such results in their entirety on the paper website, and we summarize their correlations relative to those realized by attention in a BiLSTM model with LOO measures in Figure \ref{fig:corrgl_loo}. This reports the mean differences between (i) gradient and LOO correlations, and (ii) attention and LOO correlations. As expected, we find that these exhibit, in general, considerably higher correlation with one another (on average) than LOO does with attention scores. (The lone exception is on SNLI.) Figure \ref{fig:corrgl_ag} shows the same for gradients and attention scores; the differences are comparable. In the \emph{ADR} and \emph{Diabetes} corpora, a few high precision tokens indicate (the positive) class, and in these cases we see better agreement between LOO/gradient measures with attention; this is consistent with Figure \ref{fig:AdversarialHist} which shows that it is difficult for the BiLSTM variant to find adversarial attention distributions for \emph{Diabetes}.

A potential issues with using Kendall $\tau$ as our metric here is that (potentially many) irrelevant features may add noise to the correlation measures. We acknowledge that this as a shortcoming of the metric. One observation that may mitigate this concern is that we might expect such noise to depress the LOO and gradient correlations to the same extent as they do the correlation between attention and feature importance scores; but as per Figure \ref{fig:corrgl_ag}, they do not. We also note that the correlations between the attention weights on top of feedforward (projection) encoder and LOO scores are much stronger, on average, than those between BiLSTM attention weights and LOO. This is shown in Figure \ref{fig:corral}. Were low correlations due simply to noise, we would not expect this.\footnote{The same contrast can be seen for the gradients, as one would expect given the direct gradient paths in the projection network back to individual tokens.}

The results here suggest that, in general, attention weights do not strongly or consistently agree with standard feature importance scores. The exception to this is when one uses a very simple (averaging) encoder model, as one might expect. These findings should be disconcerting to one hoping to view attention weights as explanatory, given the face validity of input gradient/erasure based explanations \cite{ross2017right,li2016understanding}.

\begin{figure*}
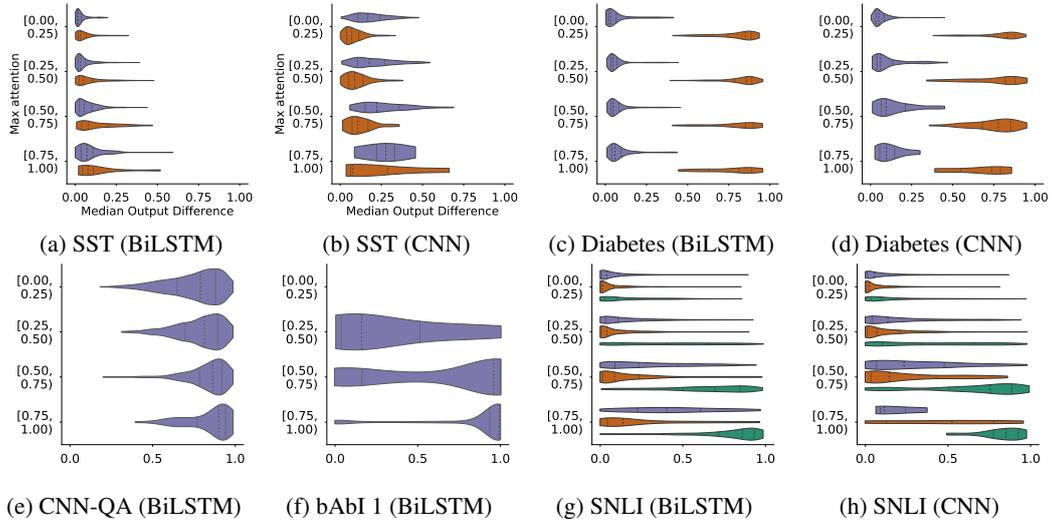

    \centering
    \AddSubFigureScatter{Permutation_MAvDY}
    \caption{\textbf{Median change in output $\boldsymbol{\Delta\hat{y}^{med}}$} (x-axis) densities in relation to the \textbf{max attention ($\boldsymbol{\max{\hat{\alpha}}}$)} (y-axis) obtained by randomly permuting instance attention weights. Encoders denoted parenthetically. Plots for all corpora and using all encoders are available online. } 
    \label{fig:PermutationScatter}
\end{figure*}

\subsection{Counterfactual Attention Weights} 
\label{section:counterfactuals}

We next consider \emph{what-if} scenarios corresponding to alternative (counterfactual) attention weights. 
The idea is to investigate whether the prediction would have been different, had the model emphasized (attended to) different input features.
More precisely, suppose $\hat{\alpha} = \{\hat{\alpha_{t}}\}_{t=1}^{T}$ are the attention weights induced for an instance, giving rise to model output $\hat{y}$. 
We then consider counterfactual distributions over $y$, under alternative $\alpha$. 
To be clear, these experiments tell us the degree to which a particular (attention) heat map uniquely induces an output. 

We experiment with two means of constructing such distributions. 
First, we simply scramble the original attention weights $\hat{\alpha}$, re-assigning each value to an arbitrary, randomly sampled index (input feature). 
Second, we generate an \emph{adversarial attention distribution}: this is a set of attention weights that is maximally distinct from $\hat{\alpha}$ but that nonetheless yields an equivalent prediction (i.e., prediction within some $\epsilon$ of $\hat{y}$).

\subsubsection{Attention Permutation}
\label{subsection:permutations}

To characterize model behavior when attention weights are shuffled, we follow Algorithm \ref{alg:permutation}.

\begin{algorithm}
\begin{algorithmic}
\State $\vect{h} \gets \textrm{Enc}(\vect{x})$, $\hat{\alpha} \gets \textrm{softmax}(\phi(\vect{h}, \vect{Q}))$
\State $\hat{y} \gets \textrm{Dec}(\vect{h}, \hat{\alpha})$ 
\For{$p \gets 1 \textrm{ to } 100$}
    \State $\alpha^p \gets \text{Permute}(\hat{\alpha})$
    \State $\hat{y}^p \gets \textrm{Dec}(\vect{h}, \alpha^p)$ \Comment{{\small Note : $\vect{h}$ is not changed}}
    \State $\Delta\hat{y}^p \gets \text{TVD}[\hat{y}^p, \hat{y}]$
\EndFor
\State $\Delta\hat{y}^{med} \gets \textrm{Median}_p(\Delta\hat{y}^p)$
\end{algorithmic}
\caption{Permuting attention weights}
\label{alg:permutation}
\end{algorithm}

\autoref{fig:PermutationScatter} depicts the relationship between the maximum attention value in the original $\hat{\alpha}$ and the median induced change in model output ($\Delta\hat{y}^{med}$) across instances in the respective datasets. Colors again indicate class predictions, as above.

We observe that there exist many points with small $\Delta\hat{y}^{med}$ despite large magnitude attention weights. These are cases in which the attention weights might suggest explaining an output by a small set of features (this is how one might reasonably read a heatmap depicting the attention weights), but where scrambling the attention makes little difference to the prediction.

\begin{figure*}
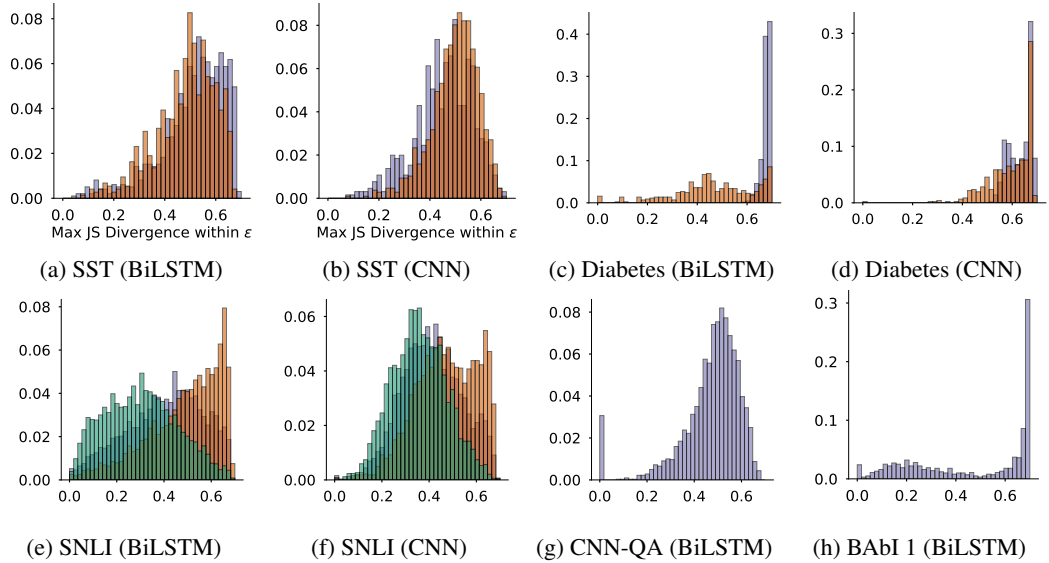

    \centering
    \AddSubFigureForAll{eMaxJDS_Hist}{cnn_1,3,5,7_}{CNN}
    \caption{Histogram of \textbf{maximum adversarial JS Divergence ($\epsilon\text{-max JSD}$)} between original and adversarial attentions over all instances. In all cases shown, $|\hat{y}^{adv} - \hat{y}| < \epsilon$. Encoders are specified in parantheses.}
    \label{fig:AdversarialHist}
\end{figure*}

In some cases, such as predicting ICD codes from notes using the MIMIC dataset, one can see different behavior for the respective classes. 
For the \emph{Diabetes} task, e.g., attention behaves intuitively for at least the positive class; perturbing attention in this case causes large changes to the prediction. 
This is consistent with our preceding observations regarding Figure \ref{fig:corrgl_ag}.
We again conjecture that this is due to a few tokens serving as high precision indicators for the positive class; in their absence (or when they are not attended to sufficiently), the prediction drops considerably.  
However, this is the exception rather than the rule. 

\subsubsection{Adversarial Attention}
\label{subsection:adversarial}

We next propose a more focused approach to counterfactual attention weights, which we will refer to as \emph{adversarial attention}. 
The intuition is to explicitly seek out attention weights that differ as much as possible from the observed attention distribution and yet leave the prediction effectively unchanged. 
Such adversarial weights violate an intuitive property of explanations: shifting model attention to very different input features should yield corresponding changes in the output. 
Alternative attention distributions identified adversarially may then be viewed as equally plausible explanations for the same output.
(A complication here is that this may be uncovering alternative, but also plausible, explanations. This would assume attention provides sufficient, but not exhaustive, rationales. This may be acceptable in some scenarios, but problematic in others.)

Operationally, realizing this objective requires specifying a value $\epsilon$ that defines what qualifies as a ``small'' difference in model output. 
Once this is specified, we aim to find $k$ adversarial distributions $\{\alpha^{(1)}, ..., \alpha^{(k)}\}$, such that each $\alpha^{(i)}$ maximizes the distance from original $\hat{\alpha}$ but does not change the output by more than $\epsilon$. In practice we simply set this to $0.01$ for text classification and $0.05$ for QA datasets.\footnote{We make the threshold slightly higher for QA because the output space is larger and thus small dimension-wise perturbations can produce comparatively large TVD.}

We propose the following optimization problem to identify adversarial attention weights. 
\begin{equation}
    \begin{aligned}
    & \underset{\alpha^{(1)}, ..., \alpha^{(k)}}{\text{maximize}} 
    & & f(\{\alpha^{(i)}\}_{i=1}^k) \\ 
    & \text{subject to}
    & & \forall i \; \text{TVD}[\hat{y}(\vect{x}, \alpha^{(i)}), \hat{y}(\vect{x},\hat{\alpha})] \le \epsilon
    \end{aligned}
    \label{eq:adv}
\end{equation}

\noindent Where $f(\{\alpha^{(i)}\}_{i=1}^k)$ is:
\begin{equation}
      \sum_{i=1}^k \text{JSD}[\alpha^{(i)}, \hat{\alpha}] + \frac{1}{k(k-1)}\sum_{i < j} \text{JSD}[\alpha^{(i)}, \alpha^{(j)}]
\end{equation}

\noindent In practice we maximize a relaxed version of this objective via the Adam SGD optimizer \cite{kingma2014adam}: $f(\{\alpha^{(i)}\}_{i=1}^k) + \frac{\lambda}{k} \sum_{i=1}^{k} \max(0, \text{TVD}[\hat{y}(\vect{x}, \alpha^{(i)}), \hat{y}(\vect{x}, \hat{\alpha})] - \epsilon)$.\footnote{We set $\lambda = 500$.}

\begin{figure*}
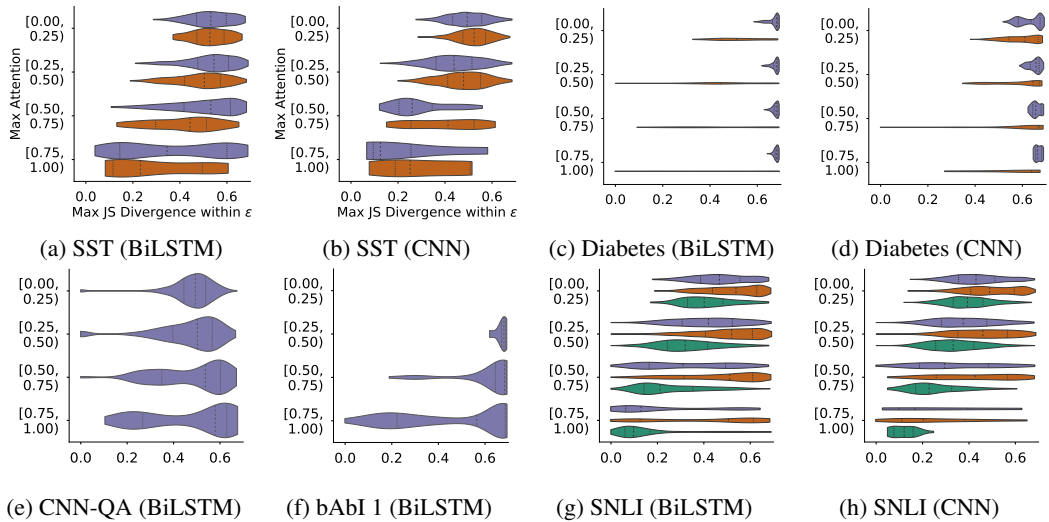

    \centering
    \AddSubFigureScatter{eMaxJDS_Scatter}
    \caption{Densities of \textbf{maximum JS divergences ($\epsilon\text{-max JSD}$)} (x-axis) as a function of the \textbf{max attention} (y-axis) in each instance for obtained between original and adversarial attention weights.}
    \label{fig:AdversarialScatter}
    \vspace{-1em}
\end{figure*}

Equation \ref{eq:adv} attempts to identify a set of new attention distributions over the input that is as far as possible from the observed $\alpha$ (as measured by JSD) and from each other (and thus diverse), while keeping the output of the model within $\epsilon$ of the original prediction. 
We denote the output obtained under the $i^{th}$ adversarial attention by $\hat{y}^{(i)}$. 
Note that the JS Divergence between any two categorical distributions (irrespective of length) is bounded from above by 0.69.

One can view an attentive decoder as a function that maps from the space of latent input representations and attention weights over input words $\Delta^{T-1}$ to a distribution over the output space $\mathcal{Y}$.
Thus, for any output $\hat{y}$, we can define how likely each attention distribution $\alpha$ will generate the output as inversely proportional to $\text{TVD}(y(\alpha), \hat{y})$.

\begin{algorithm}
\begin{algorithmic}
\State $\vect{h} \gets \textrm{Enc}(\vect{x})$, $\hat{\alpha} \gets \textrm{softmax}(\phi(\vect{h}, \vect{Q}))$
\State $\hat{y} \gets \textrm{Dec}(\vect{h}, \hat{\alpha})$
\State $\alpha^{(1)}, ..., \alpha^{(k)} \gets \text{Optimize Eq \ref{eq:adv}}$
\For{$i \gets 1 \textrm{ to } k$} 
\State $\hat{y}^{(i)} \gets \text{Dec}(\vect{h}, \alpha^{(i)})$ \Comment{{\small $\vect{h}$ is not changed}}
\State $\Delta\hat{y}^{(i)} \gets \text{TVD}[\hat{y}, \hat{y}^{(i)}]$
\State $\Delta\alpha^{(i)} \gets \text{JSD}[\hat{\alpha}, \alpha^{(i)}]$
\EndFor
\State $\epsilon\text{-max JSD} \gets \max_{i} \mathbbm{1}[\Delta\hat{y}^{(i)} \le \epsilon]\Delta\alpha^{(i)} $
\end{algorithmic}
\caption{Finding adversarial attention weights}
\label{alg:adversarial}
\end{algorithm}

\autoref{fig:AdversarialHist} depicts the distributions of max JSDs realized over instances with adversarial attention weights for a subset of the datasets considered (plots for the remaining corpora are available online). 
Colors again indicate predicted class.
Mass toward the upper-bound of 0.69 indicates that we are frequently able to identify maximally different attention weights that hardly budge model output.
We observe that one can identify adversarial attention weights associated with high JSD for a significant number of examples.
This means that it is often the case that quite different attention distributions over inputs would yield essentially the same (within $\epsilon$) output.

In the case of the \emph{Diabetes} task, we again observe a pattern of low JSD for positive examples (where evidence is present) and high JSD for negative examples. 
In other words, for this task, if one perturbs the attention weights when it is inferred that the patient is diabetic, this does change the output, which is intuitively agreeable.
However, this behavior again is an exception to the rule.

We also consider the relationship between max attention weights (indicating strong emphasis on a particular feature) and the dissimilarity of identified adversarial attention weights, as measured via JSD, for adversaries that yield a prediction within $\epsilon$ of the original model output.
Intuitively, one might hope that if attention weights are peaky, then counterfactual attention weights that are very different but which yield equivalent predictions would be more difficult to identify.

\autoref{fig:AdversarialScatter} illustrates that while there is a negative trend to this effect, it is realized only weakly.
Thus there exist many cases (in all datasets) in which despite a high attention weight, an alternative and quite different attention configuration over inputs yields effectively the same output. 
In light of this, presenting a heatmap in such a scenario that implies a particular feature is primarily responsible for an output would seem to be misleading.

\section{Related Work}
\label{section:related-work}

We have focused on attention mechanisms and the question of whether they afford transparency, but a number of interesting strategies unrelated to attention mechanisms have been recently proposed to provide insights into neural NLP models. These include approaches that measure feature importance based on gradient information \cite{ross2017right,sundararajan2017axiomatic} (aligned with the gradient-based measures that we have used here), and methods based on \emph{representation erasure} \cite{li2016understanding}, in which dimensions are removed and then the resultant change in output is recorded (similar to our experiments with removing tokens from inputs, albeit we do this at the input layer).

Comparing such importance measures to attention scores may provide additional insights into the working of attention based models \cite{ghaeini2018interpreting}. Another novel line of work in this direction involves explicitly identifying explanations of black-box predictions via a causal framework \cite{alvarez2017causal}. We also note that there has been complementary work demonstrating correlation between human attention and induced attention weights, which was relatively strong when humans agreed on an explanation \cite{pappas2016human}. It would be interesting to explore if such cases present explicit `high precision' signals in the text (for example, the positive label in diabetes dataset).

More specific to attention mechanisms, recent promising work has proposed more principled attention variants designed explicitly for interpretability; these may provide greater transparency by imposing \emph{hard}, \emph{sparse} attention. 
Such instantiations explicitly select (modest) subsets of inputs to be considered when making a prediction, which are then by construction responsible for model output \cite{lei2016rationalizing,peters2018interpretable}. 
\emph{Structured attention} models \cite{kim2017structured} provide a generalized framework for describing and fitting attention variants with explicit probabilistic semantics. Tying attention weights to human-provided rationales is another potentially promising avenue \cite{bao2018deriving}. 

We hope our work motivates further development of these methods, resulting in attention variants that both improve predictive performance and provide insights into model predictions.

\section{Discussion and Conclusions} 
\label{section:conclusions}

We have provided evidence that correlation between intuitive feature importance measures (including gradient and feature erasure approaches) and learned attention weights is weak for recurrent encoders (Section \ref{section:gradient-correlations}). We also established that counterfactual attention distributions --- which would tell a different story about why a model made the prediction that it did --- often have modest effects on model output (Section \ref{section:counterfactuals}). 

These results suggest that while attention modules consistently yield improved performance on NLP tasks, their ability to provide transparency or meaningful explanations for model predictions is, at best, questionable -- especially when a complex encoder is used, which may entangle inputs in the hidden space. 
In our view, this should not be terribly surprising, given that the encoder induces representations that may encode arbitrary interactions between individual inputs; presenting heatmaps of attention weights placed over these inputs can then be misleading.
And indeed, how one is meant to interpret such heatmaps is unclear. 
They would seem to suggest a story about how a model arrived at a particular disposition, but the results here indicate that the relationship between this and attention is not always obvious. 

There are important {\bf limitations} to this work and the conclusions we can draw from it. 
We have reported the (generally weak) correlation between learned attention weights and various alternative measures of feature importance, e.g., gradients. 
We do not intend to imply that such alternative measures are necessarily ideal or that they should be considered `ground truth'.
While such measures do enjoy a clear intrinsic (to the model) semantics, their interpretation in the context of non-linear neural networks can nonetheless be difficult for humans \cite{feng2018pathologies}.
Still, that attention consistently correlates poorly with \emph{multiple} such measures ought to give pause to practitioners. 
That said, exactly how strong such correlations `should' be in order to establish reliability as explanation is an admittedly subjective question.

We also acknowledge that irrelevant features may be contributing noise to the Kendall $\tau$ measure, thus depressing this metric artificially. However, we would highlight again that attention weights over non-recurrent encoders exhibit stronger correlations, on average, with leave-one-out (LOO) feature importance scores than do attention weights induced over BiLSTM outputs (Figure \ref{fig:corral}). Furthermore, LOO and gradient-based measures correlate with one another much more strongly than do (recurrent) attention weights and either feature importance score. It is not clear why either of these comparative relationships should hold if the poor correlation between attention and feature importance scores owes solely to noise. However, it remains a possibility that agreement is strong between attention weights and feature importance scores for the top-$k$ features only (the trouble would be defining this $k$ and then measuring correlation between non-identical sets).

An additional limitation is that we have only considered a handful of attention variants, selected to reflect common module architectures for the respective tasks included in our analysis. 
We have been particularly focused on RNNs (here, BiLSTMs) and attention, which is very commonly used. 
As we have shown, results for the simple feed-forward encoder demonstrate greater fidelity to alternative feature importance measures. 
Other feedforward models (including convolutional architectures) may enjoy similar properties, when designed with care.
Alternative attention specifications may yield different conclusions; and indeed we hope this work motivates further development of principled attention mechanisms.

It is also important to note that the counterfactual attention experiments demonstrate the existence of alternative heatmaps that yield equivalent predictions; thus one cannot conclude that the model made a particular prediction \emph{because} it attended over inputs in a specific way. However, the adversarial weights themselves may be scored as unlikely under the attention module parameters. Furthermore, it may be that multiple plausible explanations for a particular disposition exist, and this may complicate interpretation: We would maintain that in such cases the model should highlight all plausible explanations, but one may instead view a model that provides `sufficient' explanation as reasonable. 

Finally, we have limited our evaluation to tasks with unstructured output spaces, i.e., we have not considered seq2seq tasks, which we leave for future work. 
However we believe interpretability is more often a consideration in, e.g., classification than in translation. 

\section{Acknowledgements}
We thank Zachary Lipton for insightful feedback on a preliminary version of this manuscript. 

We also thank Yuval Pinter and Sarah Wiegreffe for helpful comments on previous versions of this paper. We additionally had productive discussions regarding this work with Sofia Serrano and Raymond Mooney. 

This work was supported by the Army Research Office (ARO), award W911NF1810328. 
\bibliography{naaclhlt2019}
\bibliographystyle{acl_natbib}

\clearpage 
\newpage
\onecolumn
\begin{center}
{\Large \textbf{Attention is not Explanation: Appendix}}
\end{center}
\appendix
\section{Model details}

For all datasets, we use \href{https://spacy.io/}{spaCy} for tokenization. We map out of vocabulary words to a special $\texttt{<unk>}$ token and map all words with numeric characters to `qqq'. Each word in the vocabulary was initialized to pretrained embeddings. For general domain corpora we used either (i) \href{https://s3-us-west-1.amazonaws.com/fasttext-vectors/wiki.simple.vec}{FastText Embeddings} (SST, IMDB, 20News, and CNN) trained on Simple English Wikipedia, or, (ii) \href{http://nlp.stanford.edu/data/glove.840B.300d.zip}{GloVe 840B} embeddings (AGNews and SNLI). For the MIMIC dataset, we learned word embeddings using \href{https://radimrehurek.com/gensim/}{Gensim} over all discharge summaries in the corpus. 
We initialize words not present in the vocabulary using samples from a standard Gaussian $\mathcal{N}$($\mu=0$, $\sigma^2=1$).

\subsection{BiLSTM}
We use an embedding size of 300 and hidden size of 128 for all datasets except bAbI (for which we use 50 and 30, respectively). All models were regularized using $\ell_2$ regularization ($\lambda = 10^{-5}$) applied to all parameters. We use a sigmoid activation functions for binary classification tasks, and a softmax for all other outputs. We trained the model using maximum likelihood loss using the Adam Optimizer with default parameters in PyTorch. 

\subsection{CNN}
We use an embedding size of 300 and 4 kernels of sizes [1, 3, 5, 7], each with 64 filters, giving a final hidden size of 256 (for bAbI we use 50 and 8 respectively with same kernel sizes). We use ReLU activation function on the output of the filters. All other configurations remain same as BiLSTM.

\subsection{Average}
We use the embedding size of 300 and a projection size of 256 with ReLU activation on the output of the projection matrix.
All other configurations remain same as BiLSTM.


\section{Further details regarding attentional module of gradient}

In the gradient experiments, we made the decision to cut-off the computation graph at the attention module so that gradient does not flow through this layer and contribute to the gradient feature importance score. 
For the sake of gradient calculation this effectively treats the attention as a separate input to the network, independent of the input.
We argue that this is a natural choice to make for our analysis because it calculates: \emph{how much does the output change as we perturb particular inputs (words) by a small amount, while paying the same amount of attention to said word as originally estimated and shown in the heatmap}?


\section{Graphs}

To provide easy navigation of our (large set of) graphs depicting attention weights on various datasets/tasks under various model configuration we have created an interactive interface to browse these results, accessible at: \url{https://successar.github.io/AttentionExplanation/docs/}.

\section{Adversarial Heatmaps}

\noindent\textbf{SST}

\textbf{Original}: {\setlength{\fboxsep}{0pt}\colorbox[Hsb]{202, 0.00, 1.0}{\strut reggio}} {\setlength{\fboxsep}{0pt}\colorbox[Hsb]{202, 0.01, 1.0}{\strut falls}} {\setlength{\fboxsep}{0pt}\colorbox[Hsb]{202, 0.04, 1.0}{\strut victim}} {\setlength{\fboxsep}{0pt}\colorbox[Hsb]{202, 0.00, 1.0}{\strut to}} {\setlength{\fboxsep}{0pt}\colorbox[Hsb]{202, 0.03, 1.0}{\strut relying}} {\setlength{\fboxsep}{0pt}\colorbox[Hsb]{202, 0.00, 1.0}{\strut on}} {\setlength{\fboxsep}{0pt}\colorbox[Hsb]{202, 0.00, 1.0}{\strut the}} {\setlength{\fboxsep}{0pt}\colorbox[Hsb]{202, 0.00, 1.0}{\strut very}} {\setlength{\fboxsep}{0pt}\colorbox[Hsb]{202, 0.00, 1.0}{\strut digital}} {\setlength{\fboxsep}{0pt}\colorbox[Hsb]{202, 0.00, 1.0}{\strut technology}} {\setlength{\fboxsep}{0pt}\colorbox[Hsb]{202, 0.00, 1.0}{\strut that}} {\setlength{\fboxsep}{0pt}\colorbox[Hsb]{202, 0.00, 1.0}{\strut he}} {\setlength{\fboxsep}{0pt}\colorbox[Hsb]{202, 0.00, 1.0}{\strut fervently}} {\setlength{\fboxsep}{0pt}\colorbox[Hsb]{202, 0.00, 1.0}{\strut scorns}} {\setlength{\fboxsep}{0pt}\colorbox[Hsb]{202, 0.01, 1.0}{\strut creating}} {\setlength{\fboxsep}{0pt}\colorbox[Hsb]{202, 0.00, 1.0}{\strut a}} {\setlength{\fboxsep}{0pt}\colorbox[Hsb]{202, 0.01, 1.0}{\strut meandering}} {\setlength{\fboxsep}{0pt}\colorbox[Hsb]{202, 0.00, 1.0}{\strut inarticulate}} {\setlength{\fboxsep}{0pt}\colorbox[Hsb]{202, 0.00, 1.0}{\strut and}} {\setlength{\fboxsep}{0pt}\colorbox[Hsb]{202, 0.02, 1.0}{\strut ultimately}} {\setlength{\fboxsep}{0pt}\colorbox[Hsb]{202, 0.38, 1.0}{\strut disappointing}} {\setlength{\fboxsep}{0pt}\colorbox[Hsb]{202, 0.00, 1.0}{\strut film}}

\textbf{Adversarial}: {\setlength{\fboxsep}{0pt}\colorbox[Hsb]{202, 0.00, 1.0}{\strut reggio}} {\setlength{\fboxsep}{0pt}\colorbox[Hsb]{202, 0.00, 1.0}{\strut falls}} {\setlength{\fboxsep}{0pt}\colorbox[Hsb]{202, 0.00, 1.0}{\strut victim}} {\setlength{\fboxsep}{0pt}\colorbox[Hsb]{202, 0.00, 1.0}{\strut to}} {\setlength{\fboxsep}{0pt}\colorbox[Hsb]{202, 0.00, 1.0}{\strut relying}} {\setlength{\fboxsep}{0pt}\colorbox[Hsb]{202, 0.00, 1.0}{\strut on}} {\setlength{\fboxsep}{0pt}\colorbox[Hsb]{202, 0.00, 1.0}{\strut the}} {\setlength{\fboxsep}{0pt}\colorbox[Hsb]{202, 0.00, 1.0}{\strut very}} {\setlength{\fboxsep}{0pt}\colorbox[Hsb]{202, 0.00, 1.0}{\strut digital}} {\setlength{\fboxsep}{0pt}\colorbox[Hsb]{202, 0.00, 1.0}{\strut technology}} {\setlength{\fboxsep}{0pt}\colorbox[Hsb]{202, 0.50, 1.0}{\strut that}} {\setlength{\fboxsep}{0pt}\colorbox[Hsb]{202, 0.00, 1.0}{\strut he}} {\setlength{\fboxsep}{0pt}\colorbox[Hsb]{202, 0.00, 1.0}{\strut fervently}} {\setlength{\fboxsep}{0pt}\colorbox[Hsb]{202, 0.00, 1.0}{\strut scorns}} {\setlength{\fboxsep}{0pt}\colorbox[Hsb]{202, 0.00, 1.0}{\strut creating}} {\setlength{\fboxsep}{0pt}\colorbox[Hsb]{202, 0.00, 1.0}{\strut a}} {\setlength{\fboxsep}{0pt}\colorbox[Hsb]{202, 0.00, 1.0}{\strut meandering}} {\setlength{\fboxsep}{0pt}\colorbox[Hsb]{202, 0.00, 1.0}{\strut inarticulate}} {\setlength{\fboxsep}{0pt}\colorbox[Hsb]{202, 0.00, 1.0}{\strut and}} {\setlength{\fboxsep}{0pt}\colorbox[Hsb]{202, 0.00, 1.0}{\strut ultimately}} {\setlength{\fboxsep}{0pt}\colorbox[Hsb]{202, 0.00, 1.0}{\strut disappointing}} {\setlength{\fboxsep}{0pt}\colorbox[Hsb]{202, 0.00, 1.0}{\strut film}}
\textbf{$\Delta\hat{y}$}: \emph{0.005}

\vspace{1em}
\noindent\textbf{IMDB}

\textbf{Original}: {\setlength{\fboxsep}{0pt}\colorbox[Hsb]{202, 0.09, 1.0}{\strut fantastic}} {\setlength{\fboxsep}{0pt}\colorbox[Hsb]{202, 0.00, 1.0}{\strut movie}} {\setlength{\fboxsep}{0pt}\colorbox[Hsb]{202, 0.00, 1.0}{\strut one}} {\setlength{\fboxsep}{0pt}\colorbox[Hsb]{202, 0.00, 1.0}{\strut of}} {\setlength{\fboxsep}{0pt}\colorbox[Hsb]{202, 0.00, 1.0}{\strut the}} {\setlength{\fboxsep}{0pt}\colorbox[Hsb]{202, 0.16, 1.0}{\strut best}} {\setlength{\fboxsep}{0pt}\colorbox[Hsb]{202, 0.05, 1.0}{\strut film}} {\setlength{\fboxsep}{0pt}\colorbox[Hsb]{202, 0.04, 1.0}{\strut noir}} {\setlength{\fboxsep}{0pt}\colorbox[Hsb]{202, 0.00, 1.0}{\strut movies}} {\setlength{\fboxsep}{0pt}\colorbox[Hsb]{202, 0.00, 1.0}{\strut ever}} {\setlength{\fboxsep}{0pt}\colorbox[Hsb]{202, 0.00, 1.0}{\strut made}} {\setlength{\fboxsep}{0pt}\colorbox[Hsb]{202, 0.00, 1.0}{\strut bad}} {\setlength{\fboxsep}{0pt}\colorbox[Hsb]{202, 0.00, 1.0}{\strut guys}} {\setlength{\fboxsep}{0pt}\colorbox[Hsb]{202, 0.00, 1.0}{\strut bad}} {\setlength{\fboxsep}{0pt}\colorbox[Hsb]{202, 0.00, 1.0}{\strut girls}} {\setlength{\fboxsep}{0pt}\colorbox[Hsb]{202, 0.00, 1.0}{\strut a}} {\setlength{\fboxsep}{0pt}\colorbox[Hsb]{202, 0.00, 1.0}{\strut jewel}} {\setlength{\fboxsep}{0pt}\colorbox[Hsb]{202, 0.00, 1.0}{\strut heist}} {\setlength{\fboxsep}{0pt}\colorbox[Hsb]{202, 0.00, 1.0}{\strut a}} {\setlength{\fboxsep}{0pt}\colorbox[Hsb]{202, 0.00, 1.0}{\strut twisted}} {\setlength{\fboxsep}{0pt}\colorbox[Hsb]{202, 0.00, 1.0}{\strut morality}} {\setlength{\fboxsep}{0pt}\colorbox[Hsb]{202, 0.00, 1.0}{\strut a}} {\setlength{\fboxsep}{0pt}\colorbox[Hsb]{202, 0.00, 1.0}{\strut kidnapping}} {\setlength{\fboxsep}{0pt}\colorbox[Hsb]{202, 0.00, 1.0}{\strut everything}} {\setlength{\fboxsep}{0pt}\colorbox[Hsb]{202, 0.00, 1.0}{\strut is}} {\setlength{\fboxsep}{0pt}\colorbox[Hsb]{202, 0.00, 1.0}{\strut here}} {\setlength{\fboxsep}{0pt}\colorbox[Hsb]{202, 0.00, 1.0}{\strut jean}} {\setlength{\fboxsep}{0pt}\colorbox[Hsb]{202, 0.00, 1.0}{\strut has}} {\setlength{\fboxsep}{0pt}\colorbox[Hsb]{202, 0.00, 1.0}{\strut a}} {\setlength{\fboxsep}{0pt}\colorbox[Hsb]{202, 0.00, 1.0}{\strut face}} {\setlength{\fboxsep}{0pt}\colorbox[Hsb]{202, 0.00, 1.0}{\strut that}} {\setlength{\fboxsep}{0pt}\colorbox[Hsb]{202, 0.00, 1.0}{\strut would}} {\setlength{\fboxsep}{0pt}\colorbox[Hsb]{202, 0.00, 1.0}{\strut make}} {\setlength{\fboxsep}{0pt}\colorbox[Hsb]{202, 0.00, 1.0}{\strut bogart}} {\setlength{\fboxsep}{0pt}\colorbox[Hsb]{202, 0.00, 1.0}{\strut proud}} {\setlength{\fboxsep}{0pt}\colorbox[Hsb]{202, 0.00, 1.0}{\strut and}} {\setlength{\fboxsep}{0pt}\colorbox[Hsb]{202, 0.00, 1.0}{\strut the}} {\setlength{\fboxsep}{0pt}\colorbox[Hsb]{202, 0.00, 1.0}{\strut rest}} {\setlength{\fboxsep}{0pt}\colorbox[Hsb]{202, 0.00, 1.0}{\strut of}} {\setlength{\fboxsep}{0pt}\colorbox[Hsb]{202, 0.00, 1.0}{\strut the}} {\setlength{\fboxsep}{0pt}\colorbox[Hsb]{202, 0.00, 1.0}{\strut cast}} {\setlength{\fboxsep}{0pt}\colorbox[Hsb]{202, 0.00, 1.0}{\strut is}} {\setlength{\fboxsep}{0pt}\colorbox[Hsb]{202, 0.00, 1.0}{\strut is}} {\setlength{\fboxsep}{0pt}\colorbox[Hsb]{202, 0.00, 1.0}{\strut full}} {\setlength{\fboxsep}{0pt}\colorbox[Hsb]{202, 0.00, 1.0}{\strut of}} {\setlength{\fboxsep}{0pt}\colorbox[Hsb]{202, 0.00, 1.0}{\strut character}} {\setlength{\fboxsep}{0pt}\colorbox[Hsb]{202, 0.00, 1.0}{\strut actors}} {\setlength{\fboxsep}{0pt}\colorbox[Hsb]{202, 0.00, 1.0}{\strut who}} {\setlength{\fboxsep}{0pt}\colorbox[Hsb]{202, 0.00, 1.0}{\strut seem}} {\setlength{\fboxsep}{0pt}\colorbox[Hsb]{202, 0.00, 1.0}{\strut to}} {\setlength{\fboxsep}{0pt}\colorbox[Hsb]{202, 0.00, 1.0}{\strut to}} {\setlength{\fboxsep}{0pt}\colorbox[Hsb]{202, 0.00, 1.0}{\strut know}} {\setlength{\fboxsep}{0pt}\colorbox[Hsb]{202, 0.00, 1.0}{\strut they're}} {\setlength{\fboxsep}{0pt}\colorbox[Hsb]{202, 0.00, 1.0}{\strut onto}} {\setlength{\fboxsep}{0pt}\colorbox[Hsb]{202, 0.00, 1.0}{\strut something}} {\setlength{\fboxsep}{0pt}\colorbox[Hsb]{202, 0.00, 1.0}{\strut good}} {\setlength{\fboxsep}{0pt}\colorbox[Hsb]{202, 0.00, 1.0}{\strut get}} {\setlength{\fboxsep}{0pt}\colorbox[Hsb]{202, 0.00, 1.0}{\strut some}} {\setlength{\fboxsep}{0pt}\colorbox[Hsb]{202, 0.00, 1.0}{\strut popcorn}} {\setlength{\fboxsep}{0pt}\colorbox[Hsb]{202, 0.00, 1.0}{\strut and}} {\setlength{\fboxsep}{0pt}\colorbox[Hsb]{202, 0.00, 1.0}{\strut have}} {\setlength{\fboxsep}{0pt}\colorbox[Hsb]{202, 0.00, 1.0}{\strut a}} {\setlength{\fboxsep}{0pt}\colorbox[Hsb]{202, 0.11, 1.0}{\strut great}} {\setlength{\fboxsep}{0pt}\colorbox[Hsb]{202, 0.03, 1.0}{\strut time}}

\textbf{Adversarial}: {\setlength{\fboxsep}{0pt}\colorbox[Hsb]{202, 0.00, 1.0}{\strut fantastic}} {\setlength{\fboxsep}{0pt}\colorbox[Hsb]{202, 0.00, 1.0}{\strut movie}} {\setlength{\fboxsep}{0pt}\colorbox[Hsb]{202, 0.00, 1.0}{\strut one}} {\setlength{\fboxsep}{0pt}\colorbox[Hsb]{202, 0.00, 1.0}{\strut of}} {\setlength{\fboxsep}{0pt}\colorbox[Hsb]{202, 0.00, 1.0}{\strut the}} {\setlength{\fboxsep}{0pt}\colorbox[Hsb]{202, 0.00, 1.0}{\strut best}} {\setlength{\fboxsep}{0pt}\colorbox[Hsb]{202, 0.00, 1.0}{\strut film}} {\setlength{\fboxsep}{0pt}\colorbox[Hsb]{202, 0.00, 1.0}{\strut noir}} {\setlength{\fboxsep}{0pt}\colorbox[Hsb]{202, 0.00, 1.0}{\strut movies}} {\setlength{\fboxsep}{0pt}\colorbox[Hsb]{202, 0.00, 1.0}{\strut ever}} {\setlength{\fboxsep}{0pt}\colorbox[Hsb]{202, 0.00, 1.0}{\strut made}} {\setlength{\fboxsep}{0pt}\colorbox[Hsb]{202, 0.00, 1.0}{\strut bad}} {\setlength{\fboxsep}{0pt}\colorbox[Hsb]{202, 0.00, 1.0}{\strut guys}} {\setlength{\fboxsep}{0pt}\colorbox[Hsb]{202, 0.00, 1.0}{\strut bad}} {\setlength{\fboxsep}{0pt}\colorbox[Hsb]{202, 0.00, 1.0}{\strut girls}} {\setlength{\fboxsep}{0pt}\colorbox[Hsb]{202, 0.00, 1.0}{\strut a}} {\setlength{\fboxsep}{0pt}\colorbox[Hsb]{202, 0.00, 1.0}{\strut jewel}} {\setlength{\fboxsep}{0pt}\colorbox[Hsb]{202, 0.00, 1.0}{\strut heist}} {\setlength{\fboxsep}{0pt}\colorbox[Hsb]{202, 0.00, 1.0}{\strut a}} {\setlength{\fboxsep}{0pt}\colorbox[Hsb]{202, 0.00, 1.0}{\strut twisted}} {\setlength{\fboxsep}{0pt}\colorbox[Hsb]{202, 0.00, 1.0}{\strut morality}} {\setlength{\fboxsep}{0pt}\colorbox[Hsb]{202, 0.00, 1.0}{\strut a}} {\setlength{\fboxsep}{0pt}\colorbox[Hsb]{202, 0.00, 1.0}{\strut kidnapping}} {\setlength{\fboxsep}{0pt}\colorbox[Hsb]{202, 0.50, 1.0}{\strut everything}} {\setlength{\fboxsep}{0pt}\colorbox[Hsb]{202, 0.00, 1.0}{\strut is}} {\setlength{\fboxsep}{0pt}\colorbox[Hsb]{202, 0.00, 1.0}{\strut here}} {\setlength{\fboxsep}{0pt}\colorbox[Hsb]{202, 0.00, 1.0}{\strut jean}} {\setlength{\fboxsep}{0pt}\colorbox[Hsb]{202, 0.00, 1.0}{\strut has}} {\setlength{\fboxsep}{0pt}\colorbox[Hsb]{202, 0.00, 1.0}{\strut a}} {\setlength{\fboxsep}{0pt}\colorbox[Hsb]{202, 0.00, 1.0}{\strut face}} {\setlength{\fboxsep}{0pt}\colorbox[Hsb]{202, 0.00, 1.0}{\strut that}} {\setlength{\fboxsep}{0pt}\colorbox[Hsb]{202, 0.00, 1.0}{\strut would}} {\setlength{\fboxsep}{0pt}\colorbox[Hsb]{202, 0.00, 1.0}{\strut make}} {\setlength{\fboxsep}{0pt}\colorbox[Hsb]{202, 0.00, 1.0}{\strut bogart}} {\setlength{\fboxsep}{0pt}\colorbox[Hsb]{202, 0.00, 1.0}{\strut proud}} {\setlength{\fboxsep}{0pt}\colorbox[Hsb]{202, 0.00, 1.0}{\strut and}} {\setlength{\fboxsep}{0pt}\colorbox[Hsb]{202, 0.00, 1.0}{\strut the}} {\setlength{\fboxsep}{0pt}\colorbox[Hsb]{202, 0.00, 1.0}{\strut rest}} {\setlength{\fboxsep}{0pt}\colorbox[Hsb]{202, 0.00, 1.0}{\strut of}} {\setlength{\fboxsep}{0pt}\colorbox[Hsb]{202, 0.00, 1.0}{\strut the}} {\setlength{\fboxsep}{0pt}\colorbox[Hsb]{202, 0.00, 1.0}{\strut cast}} {\setlength{\fboxsep}{0pt}\colorbox[Hsb]{202, 0.00, 1.0}{\strut is}} {\setlength{\fboxsep}{0pt}\colorbox[Hsb]{202, 0.00, 1.0}{\strut is}} {\setlength{\fboxsep}{0pt}\colorbox[Hsb]{202, 0.00, 1.0}{\strut full}} {\setlength{\fboxsep}{0pt}\colorbox[Hsb]{202, 0.00, 1.0}{\strut of}} {\setlength{\fboxsep}{0pt}\colorbox[Hsb]{202, 0.00, 1.0}{\strut character}} {\setlength{\fboxsep}{0pt}\colorbox[Hsb]{202, 0.00, 1.0}{\strut actors}} {\setlength{\fboxsep}{0pt}\colorbox[Hsb]{202, 0.00, 1.0}{\strut who}} {\setlength{\fboxsep}{0pt}\colorbox[Hsb]{202, 0.00, 1.0}{\strut seem}} {\setlength{\fboxsep}{0pt}\colorbox[Hsb]{202, 0.00, 1.0}{\strut to}} {\setlength{\fboxsep}{0pt}\colorbox[Hsb]{202, 0.00, 1.0}{\strut to}} {\setlength{\fboxsep}{0pt}\colorbox[Hsb]{202, 0.00, 1.0}{\strut know}} {\setlength{\fboxsep}{0pt}\colorbox[Hsb]{202, 0.00, 1.0}{\strut they're}} {\setlength{\fboxsep}{0pt}\colorbox[Hsb]{202, 0.00, 1.0}{\strut onto}} {\setlength{\fboxsep}{0pt}\colorbox[Hsb]{202, 0.00, 1.0}{\strut something}} {\setlength{\fboxsep}{0pt}\colorbox[Hsb]{202, 0.00, 1.0}{\strut good}} {\setlength{\fboxsep}{0pt}\colorbox[Hsb]{202, 0.00, 1.0}{\strut get}} {\setlength{\fboxsep}{0pt}\colorbox[Hsb]{202, 0.00, 1.0}{\strut some}} {\setlength{\fboxsep}{0pt}\colorbox[Hsb]{202, 0.00, 1.0}{\strut popcorn}} {\setlength{\fboxsep}{0pt}\colorbox[Hsb]{202, 0.00, 1.0}{\strut and}} {\setlength{\fboxsep}{0pt}\colorbox[Hsb]{202, 0.00, 1.0}{\strut have}} {\setlength{\fboxsep}{0pt}\colorbox[Hsb]{202, 0.00, 1.0}{\strut a}} {\setlength{\fboxsep}{0pt}\colorbox[Hsb]{202, 0.00, 1.0}{\strut great}} {\setlength{\fboxsep}{0pt}\colorbox[Hsb]{202, 0.00, 1.0}{\strut time}} 
\textbf{$\Delta\hat{y}$}: \emph{0.004}

\vspace{1em}
\par
\noindent\textbf{20 News Group - Sports}

\textbf{Original}:{\setlength{\fboxsep}{0pt}\colorbox[Hsb]{202, 0.00, 1.0}{\strut i}} {\setlength{\fboxsep}{0pt}\colorbox[Hsb]{202, 0.00, 1.0}{\strut meant}} {\setlength{\fboxsep}{0pt}\colorbox[Hsb]{202, 0.00, 1.0}{\strut to}} {\setlength{\fboxsep}{0pt}\colorbox[Hsb]{202, 0.00, 1.0}{\strut comment}} {\setlength{\fboxsep}{0pt}\colorbox[Hsb]{202, 0.00, 1.0}{\strut on}} {\setlength{\fboxsep}{0pt}\colorbox[Hsb]{202, 0.00, 1.0}{\strut this}} {\setlength{\fboxsep}{0pt}\colorbox[Hsb]{202, 0.00, 1.0}{\strut at}} {\setlength{\fboxsep}{0pt}\colorbox[Hsb]{202, 0.00, 1.0}{\strut the}} {\setlength{\fboxsep}{0pt}\colorbox[Hsb]{202, 0.00, 1.0}{\strut time}} {\setlength{\fboxsep}{0pt}\colorbox[Hsb]{202, 0.00, 1.0}{\strut there}} {\setlength{\fboxsep}{0pt}\colorbox[Hsb]{202, 0.00, 1.0}{\strut '}} {\setlength{\fboxsep}{0pt}\colorbox[Hsb]{202, 0.00, 1.0}{\strut s}} {\setlength{\fboxsep}{0pt}\colorbox[Hsb]{202, 0.00, 1.0}{\strut just}} {\setlength{\fboxsep}{0pt}\colorbox[Hsb]{202, 0.00, 1.0}{\strut no}} {\setlength{\fboxsep}{0pt}\colorbox[Hsb]{202, 0.00, 1.0}{\strut way}} {\setlength{\fboxsep}{0pt}\colorbox[Hsb]{202, 0.00, 1.0}{\strut baserunning}} {\setlength{\fboxsep}{0pt}\colorbox[Hsb]{202, 0.00, 1.0}{\strut could}} {\setlength{\fboxsep}{0pt}\colorbox[Hsb]{202, 0.00, 1.0}{\strut be}} {\setlength{\fboxsep}{0pt}\colorbox[Hsb]{202, 0.00, 1.0}{\strut that}} {\setlength{\fboxsep}{0pt}\colorbox[Hsb]{202, 0.00, 1.0}{\strut important}} {\setlength{\fboxsep}{0pt}\colorbox[Hsb]{202, 0.00, 1.0}{\strut if}} {\setlength{\fboxsep}{0pt}\colorbox[Hsb]{202, 0.00, 1.0}{\strut it}} {\setlength{\fboxsep}{0pt}\colorbox[Hsb]{202, 0.00, 1.0}{\strut was}} {\setlength{\fboxsep}{0pt}\colorbox[Hsb]{202, 0.00, 1.0}{\strut runs}} {\setlength{\fboxsep}{0pt}\colorbox[Hsb]{202, 0.01, 1.0}{\strut created}} {\setlength{\fboxsep}{0pt}\colorbox[Hsb]{202, 0.00, 1.0}{\strut would}} {\setlength{\fboxsep}{0pt}\colorbox[Hsb]{202, 0.00, 1.0}{\strut n}} {\setlength{\fboxsep}{0pt}\colorbox[Hsb]{202, 0.00, 1.0}{\strut '}} {\setlength{\fboxsep}{0pt}\colorbox[Hsb]{202, 0.00, 1.0}{\strut t}} {\setlength{\fboxsep}{0pt}\colorbox[Hsb]{202, 0.00, 1.0}{\strut be}} {\setlength{\fboxsep}{0pt}\colorbox[Hsb]{202, 0.00, 1.0}{\strut nearly}} {\setlength{\fboxsep}{0pt}\colorbox[Hsb]{202, 0.00, 1.0}{\strut as}} {\setlength{\fboxsep}{0pt}\colorbox[Hsb]{202, 0.00, 1.0}{\strut accurate}} {\setlength{\fboxsep}{0pt}\colorbox[Hsb]{202, 0.00, 1.0}{\strut as}} {\setlength{\fboxsep}{0pt}\colorbox[Hsb]{202, 0.00, 1.0}{\strut it}} {\setlength{\fboxsep}{0pt}\colorbox[Hsb]{202, 0.00, 1.0}{\strut is}} {\setlength{\fboxsep}{0pt}\colorbox[Hsb]{202, 0.00, 1.0}{\strut runs}} {\setlength{\fboxsep}{0pt}\colorbox[Hsb]{202, 0.00, 1.0}{\strut created}} {\setlength{\fboxsep}{0pt}\colorbox[Hsb]{202, 0.00, 1.0}{\strut is}} {\setlength{\fboxsep}{0pt}\colorbox[Hsb]{202, 0.01, 1.0}{\strut usually}} {\setlength{\fboxsep}{0pt}\colorbox[Hsb]{202, 0.01, 1.0}{\strut about}} {\setlength{\fboxsep}{0pt}\colorbox[Hsb]{202, 0.01, 1.0}{\strut qqq}} {\setlength{\fboxsep}{0pt}\colorbox[Hsb]{202, 0.02, 1.0}{\strut qqq}} {\setlength{\fboxsep}{0pt}\colorbox[Hsb]{202, 0.02, 1.0}{\strut accurate}} {\setlength{\fboxsep}{0pt}\colorbox[Hsb]{202, 0.02, 1.0}{\strut on}} {\setlength{\fboxsep}{0pt}\colorbox[Hsb]{202, 0.02, 1.0}{\strut a}} {\setlength{\fboxsep}{0pt}\colorbox[Hsb]{202, 0.01, 1.0}{\strut team}} {\setlength{\fboxsep}{0pt}\colorbox[Hsb]{202, 0.02, 1.0}{\strut level}} {\setlength{\fboxsep}{0pt}\colorbox[Hsb]{202, 0.02, 1.0}{\strut and}} {\setlength{\fboxsep}{0pt}\colorbox[Hsb]{202, 0.01, 1.0}{\strut there}} {\setlength{\fboxsep}{0pt}\colorbox[Hsb]{202, 0.01, 1.0}{\strut '}} {\setlength{\fboxsep}{0pt}\colorbox[Hsb]{202, 0.01, 1.0}{\strut s}} {\setlength{\fboxsep}{0pt}\colorbox[Hsb]{202, 0.01, 1.0}{\strut a}} {\setlength{\fboxsep}{0pt}\colorbox[Hsb]{202, 0.01, 1.0}{\strut lot}} {\setlength{\fboxsep}{0pt}\colorbox[Hsb]{202, 0.01, 1.0}{\strut more}} {\setlength{\fboxsep}{0pt}\colorbox[Hsb]{202, 0.01, 1.0}{\strut than}} {\setlength{\fboxsep}{0pt}\colorbox[Hsb]{202, 0.03, 1.0}{\strut baserunning}} {\setlength{\fboxsep}{0pt}\colorbox[Hsb]{202, 0.02, 1.0}{\strut that}} {\setlength{\fboxsep}{0pt}\colorbox[Hsb]{202, 0.02, 1.0}{\strut has}} {\setlength{\fboxsep}{0pt}\colorbox[Hsb]{202, 0.02, 1.0}{\strut to}} {\setlength{\fboxsep}{0pt}\colorbox[Hsb]{202, 0.01, 1.0}{\strut account}} {\setlength{\fboxsep}{0pt}\colorbox[Hsb]{202, 0.02, 1.0}{\strut for}} {\setlength{\fboxsep}{0pt}\colorbox[Hsb]{202, 0.03, 1.0}{\strut the}} {\setlength{\fboxsep}{0pt}\colorbox[Hsb]{202, 0.03, 1.0}{\strut remaining}} {\setlength{\fboxsep}{0pt}\colorbox[Hsb]{202, 0.03, 1.0}{\strut percent}} {\setlength{\fboxsep}{0pt}\colorbox[Hsb]{202, 0.05, 1.0}{\strut .}}

\textbf{Adversarial}:{\setlength{\fboxsep}{0pt}\colorbox[Hsb]{202, 0.00, 1.0}{\strut i}} {\setlength{\fboxsep}{0pt}\colorbox[Hsb]{202, 0.00, 1.0}{\strut meant}} {\setlength{\fboxsep}{0pt}\colorbox[Hsb]{202, 0.00, 1.0}{\strut to}} {\setlength{\fboxsep}{0pt}\colorbox[Hsb]{202, 0.00, 1.0}{\strut comment}} {\setlength{\fboxsep}{0pt}\colorbox[Hsb]{202, 0.00, 1.0}{\strut on}} {\setlength{\fboxsep}{0pt}\colorbox[Hsb]{202, 0.00, 1.0}{\strut this}} {\setlength{\fboxsep}{0pt}\colorbox[Hsb]{202, 0.00, 1.0}{\strut at}} {\setlength{\fboxsep}{0pt}\colorbox[Hsb]{202, 0.00, 1.0}{\strut the}} {\setlength{\fboxsep}{0pt}\colorbox[Hsb]{202, 0.00, 1.0}{\strut time}} {\setlength{\fboxsep}{0pt}\colorbox[Hsb]{202, 0.00, 1.0}{\strut there}} {\setlength{\fboxsep}{0pt}\colorbox[Hsb]{202, 0.00, 1.0}{\strut '}} {\setlength{\fboxsep}{0pt}\colorbox[Hsb]{202, 0.00, 1.0}{\strut s}} {\setlength{\fboxsep}{0pt}\colorbox[Hsb]{202, 0.00, 1.0}{\strut just}} {\setlength{\fboxsep}{0pt}\colorbox[Hsb]{202, 0.00, 1.0}{\strut no}} {\setlength{\fboxsep}{0pt}\colorbox[Hsb]{202, 0.00, 1.0}{\strut way}} {\setlength{\fboxsep}{0pt}\colorbox[Hsb]{202, 0.00, 1.0}{\strut baserunning}} {\setlength{\fboxsep}{0pt}\colorbox[Hsb]{202, 0.00, 1.0}{\strut could}} {\setlength{\fboxsep}{0pt}\colorbox[Hsb]{202, 0.00, 1.0}{\strut be}} {\setlength{\fboxsep}{0pt}\colorbox[Hsb]{202, 0.00, 1.0}{\strut that}} {\setlength{\fboxsep}{0pt}\colorbox[Hsb]{202, 0.00, 1.0}{\strut important}} {\setlength{\fboxsep}{0pt}\colorbox[Hsb]{202, 0.00, 1.0}{\strut if}} {\setlength{\fboxsep}{0pt}\colorbox[Hsb]{202, 0.00, 1.0}{\strut it}} {\setlength{\fboxsep}{0pt}\colorbox[Hsb]{202, 0.00, 1.0}{\strut was}} {\setlength{\fboxsep}{0pt}\colorbox[Hsb]{202, 0.50, 1.0}{\strut runs}} {\setlength{\fboxsep}{0pt}\colorbox[Hsb]{202, 0.00, 1.0}{\strut created}} {\setlength{\fboxsep}{0pt}\colorbox[Hsb]{202, 0.00, 1.0}{\strut would}} {\setlength{\fboxsep}{0pt}\colorbox[Hsb]{202, 0.00, 1.0}{\strut n}} {\setlength{\fboxsep}{0pt}\colorbox[Hsb]{202, 0.00, 1.0}{\strut '}} {\setlength{\fboxsep}{0pt}\colorbox[Hsb]{202, 0.00, 1.0}{\strut t}} {\setlength{\fboxsep}{0pt}\colorbox[Hsb]{202, 0.00, 1.0}{\strut be}} {\setlength{\fboxsep}{0pt}\colorbox[Hsb]{202, 0.00, 1.0}{\strut nearly}} {\setlength{\fboxsep}{0pt}\colorbox[Hsb]{202, 0.00, 1.0}{\strut as}} {\setlength{\fboxsep}{0pt}\colorbox[Hsb]{202, 0.00, 1.0}{\strut accurate}} {\setlength{\fboxsep}{0pt}\colorbox[Hsb]{202, 0.00, 1.0}{\strut as}} {\setlength{\fboxsep}{0pt}\colorbox[Hsb]{202, 0.00, 1.0}{\strut it}} {\setlength{\fboxsep}{0pt}\colorbox[Hsb]{202, 0.00, 1.0}{\strut is}} {\setlength{\fboxsep}{0pt}\colorbox[Hsb]{202, 0.00, 1.0}{\strut runs}} {\setlength{\fboxsep}{0pt}\colorbox[Hsb]{202, 0.00, 1.0}{\strut created}} {\setlength{\fboxsep}{0pt}\colorbox[Hsb]{202, 0.00, 1.0}{\strut is}} {\setlength{\fboxsep}{0pt}\colorbox[Hsb]{202, 0.00, 1.0}{\strut usually}} {\setlength{\fboxsep}{0pt}\colorbox[Hsb]{202, 0.00, 1.0}{\strut about}} {\setlength{\fboxsep}{0pt}\colorbox[Hsb]{202, 0.00, 1.0}{\strut qqq}} {\setlength{\fboxsep}{0pt}\colorbox[Hsb]{202, 0.00, 1.0}{\strut qqq}} {\setlength{\fboxsep}{0pt}\colorbox[Hsb]{202, 0.00, 1.0}{\strut accurate}} {\setlength{\fboxsep}{0pt}\colorbox[Hsb]{202, 0.00, 1.0}{\strut on}} {\setlength{\fboxsep}{0pt}\colorbox[Hsb]{202, 0.00, 1.0}{\strut a}} {\setlength{\fboxsep}{0pt}\colorbox[Hsb]{202, 0.00, 1.0}{\strut team}} {\setlength{\fboxsep}{0pt}\colorbox[Hsb]{202, 0.00, 1.0}{\strut level}} {\setlength{\fboxsep}{0pt}\colorbox[Hsb]{202, 0.00, 1.0}{\strut and}} {\setlength{\fboxsep}{0pt}\colorbox[Hsb]{202, 0.00, 1.0}{\strut there}} {\setlength{\fboxsep}{0pt}\colorbox[Hsb]{202, 0.00, 1.0}{\strut '}} {\setlength{\fboxsep}{0pt}\colorbox[Hsb]{202, 0.00, 1.0}{\strut s}} {\setlength{\fboxsep}{0pt}\colorbox[Hsb]{202, 0.00, 1.0}{\strut a}} {\setlength{\fboxsep}{0pt}\colorbox[Hsb]{202, 0.00, 1.0}{\strut lot}} {\setlength{\fboxsep}{0pt}\colorbox[Hsb]{202, 0.00, 1.0}{\strut more}} {\setlength{\fboxsep}{0pt}\colorbox[Hsb]{202, 0.00, 1.0}{\strut than}} {\setlength{\fboxsep}{0pt}\colorbox[Hsb]{202, 0.00, 1.0}{\strut baserunning}} {\setlength{\fboxsep}{0pt}\colorbox[Hsb]{202, 0.00, 1.0}{\strut that}} {\setlength{\fboxsep}{0pt}\colorbox[Hsb]{202, 0.00, 1.0}{\strut has}} {\setlength{\fboxsep}{0pt}\colorbox[Hsb]{202, 0.00, 1.0}{\strut to}} {\setlength{\fboxsep}{0pt}\colorbox[Hsb]{202, 0.00, 1.0}{\strut account}} {\setlength{\fboxsep}{0pt}\colorbox[Hsb]{202, 0.00, 1.0}{\strut for}} {\setlength{\fboxsep}{0pt}\colorbox[Hsb]{202, 0.00, 1.0}{\strut the}} {\setlength{\fboxsep}{0pt}\colorbox[Hsb]{202, 0.00, 1.0}{\strut remaining}} {\setlength{\fboxsep}{0pt}\colorbox[Hsb]{202, 0.00, 1.0}{\strut percent}} {\setlength{\fboxsep}{0pt}\colorbox[Hsb]{202, 0.00, 1.0}{\strut .}}
\textbf{$\Delta\hat{y}$}: \emph{0.001}

\vspace{1em}
\par
\noindent\textbf{ADR}

\textbf{Original}:{\setlength{\fboxsep}{0pt}\colorbox[Hsb]{202, 0.00, 1.0}{\strut meanwhile}} {\setlength{\fboxsep}{0pt}\colorbox[Hsb]{202, 0.00, 1.0}{\strut wait}} {\setlength{\fboxsep}{0pt}\colorbox[Hsb]{202, 0.00, 1.0}{\strut for}} {\setlength{\fboxsep}{0pt}\colorbox[Hsb]{202, 0.00, 1.0}{\strut DRUG}} {\setlength{\fboxsep}{0pt}\colorbox[Hsb]{202, 0.01, 1.0}{\strut and}} {\setlength{\fboxsep}{0pt}\colorbox[Hsb]{202, 0.00, 1.0}{\strut DRUG}} {\setlength{\fboxsep}{0pt}\colorbox[Hsb]{202, 0.01, 1.0}{\strut to}} {\setlength{\fboxsep}{0pt}\colorbox[Hsb]{202, 0.15, 1.0}{\strut kick}} {\setlength{\fboxsep}{0pt}\colorbox[Hsb]{202, 0.00, 1.0}{\strut in}} {\setlength{\fboxsep}{0pt}\colorbox[Hsb]{202, 0.00, 1.0}{\strut first}} {\setlength{\fboxsep}{0pt}\colorbox[Hsb]{202, 0.00, 1.0}{\strut co}} {\setlength{\fboxsep}{0pt}\colorbox[Hsb]{202, 0.00, 1.0}{\strut i}} {\setlength{\fboxsep}{0pt}\colorbox[Hsb]{202, 0.00, 1.0}{\strut need}} {\setlength{\fboxsep}{0pt}\colorbox[Hsb]{202, 0.00, 1.0}{\strut to}} {\setlength{\fboxsep}{0pt}\colorbox[Hsb]{202, 0.01, 1.0}{\strut prep}} {\setlength{\fboxsep}{0pt}\colorbox[Hsb]{202, 0.15, 1.0}{\strut dog}} {\setlength{\fboxsep}{0pt}\colorbox[Hsb]{202, 0.16, 1.0}{\strut food}} {\setlength{\fboxsep}{0pt}\colorbox[Hsb]{202, 0.01, 1.0}{\strut etc}} {\setlength{\fboxsep}{0pt}\colorbox[Hsb]{202, 0.00, 1.0}{\strut .}} {\setlength{\fboxsep}{0pt}\colorbox[Hsb]{202, 0.00, 1.0}{\strut co}} {\setlength{\fboxsep}{0pt}\colorbox[Hsb]{202, 0.00, 1.0}{\strut omg}} {\setlength{\fboxsep}{0pt}\colorbox[Hsb]{202, 0.00, 1.0}{\strut $<$UNK$>$}} {\setlength{\fboxsep}{0pt}\colorbox[Hsb]{202, 0.00, 1.0}{\strut .}}

\textbf{Adversarial}:{\setlength{\fboxsep}{0pt}\colorbox[Hsb]{202, 0.00, 1.0}{\strut meanwhile}} {\setlength{\fboxsep}{0pt}\colorbox[Hsb]{202, 0.00, 1.0}{\strut wait}} {\setlength{\fboxsep}{0pt}\colorbox[Hsb]{202, 0.00, 1.0}{\strut for}} {\setlength{\fboxsep}{0pt}\colorbox[Hsb]{202, 0.00, 1.0}{\strut DRUG}} {\setlength{\fboxsep}{0pt}\colorbox[Hsb]{202, 0.00, 1.0}{\strut and}} {\setlength{\fboxsep}{0pt}\colorbox[Hsb]{202, 0.00, 1.0}{\strut DRUG}} {\setlength{\fboxsep}{0pt}\colorbox[Hsb]{202, 0.00, 1.0}{\strut to}} {\setlength{\fboxsep}{0pt}\colorbox[Hsb]{202, 0.00, 1.0}{\strut kick}} {\setlength{\fboxsep}{0pt}\colorbox[Hsb]{202, 0.00, 1.0}{\strut in}} {\setlength{\fboxsep}{0pt}\colorbox[Hsb]{202, 0.50, 1.0}{\strut first}} {\setlength{\fboxsep}{0pt}\colorbox[Hsb]{202, 0.00, 1.0}{\strut co}} {\setlength{\fboxsep}{0pt}\colorbox[Hsb]{202, 0.00, 1.0}{\strut i}} {\setlength{\fboxsep}{0pt}\colorbox[Hsb]{202, 0.00, 1.0}{\strut need}} {\setlength{\fboxsep}{0pt}\colorbox[Hsb]{202, 0.00, 1.0}{\strut to}} {\setlength{\fboxsep}{0pt}\colorbox[Hsb]{202, 0.00, 1.0}{\strut prep}} {\setlength{\fboxsep}{0pt}\colorbox[Hsb]{202, 0.00, 1.0}{\strut dog}} {\setlength{\fboxsep}{0pt}\colorbox[Hsb]{202, 0.00, 1.0}{\strut food}} {\setlength{\fboxsep}{0pt}\colorbox[Hsb]{202, 0.00, 1.0}{\strut etc}} {\setlength{\fboxsep}{0pt}\colorbox[Hsb]{202, 0.00, 1.0}{\strut .}} {\setlength{\fboxsep}{0pt}\colorbox[Hsb]{202, 0.00, 1.0}{\strut co}} {\setlength{\fboxsep}{0pt}\colorbox[Hsb]{202, 0.00, 1.0}{\strut omg}} {\setlength{\fboxsep}{0pt}\colorbox[Hsb]{202, 0.00, 1.0}{\strut $<$UNK$>$}} {\setlength{\fboxsep}{0pt}\colorbox[Hsb]{202, 0.00, 1.0}{\strut .}}
\textbf{$\Delta\hat{y}$}: \emph{0.002}

\vspace{1em}
\par
\noindent\textbf{AG News}

\textbf{Original}:{\setlength{\fboxsep}{0pt}\colorbox[Hsb]{202, 0.01, 1.0}{\strut general}} {\setlength{\fboxsep}{0pt}\colorbox[Hsb]{202, 0.25, 1.0}{\strut motors}} {\setlength{\fboxsep}{0pt}\colorbox[Hsb]{202, 0.00, 1.0}{\strut and}} {\setlength{\fboxsep}{0pt}\colorbox[Hsb]{202, 0.15, 1.0}{\strut daimlerchrysler}} {\setlength{\fboxsep}{0pt}\colorbox[Hsb]{202, 0.00, 1.0}{\strut say}} {\setlength{\fboxsep}{0pt}\colorbox[Hsb]{202, 0.00, 1.0}{\strut they}} {\setlength{\fboxsep}{0pt}\colorbox[Hsb]{202, 0.00, 1.0}{\strut \#}} {\setlength{\fboxsep}{0pt}\colorbox[Hsb]{202, 0.00, 1.0}{\strut qqq}} {\setlength{\fboxsep}{0pt}\colorbox[Hsb]{202, 0.00, 1.0}{\strut teaming}} {\setlength{\fboxsep}{0pt}\colorbox[Hsb]{202, 0.00, 1.0}{\strut up}} {\setlength{\fboxsep}{0pt}\colorbox[Hsb]{202, 0.00, 1.0}{\strut to}} {\setlength{\fboxsep}{0pt}\colorbox[Hsb]{202, 0.00, 1.0}{\strut develop}} {\setlength{\fboxsep}{0pt}\colorbox[Hsb]{202, 0.00, 1.0}{\strut hybrid}} {\setlength{\fboxsep}{0pt}\colorbox[Hsb]{202, 0.01, 1.0}{\strut technology}} {\setlength{\fboxsep}{0pt}\colorbox[Hsb]{202, 0.00, 1.0}{\strut for}} {\setlength{\fboxsep}{0pt}\colorbox[Hsb]{202, 0.00, 1.0}{\strut use}} {\setlength{\fboxsep}{0pt}\colorbox[Hsb]{202, 0.00, 1.0}{\strut in}} {\setlength{\fboxsep}{0pt}\colorbox[Hsb]{202, 0.00, 1.0}{\strut their}} {\setlength{\fboxsep}{0pt}\colorbox[Hsb]{202, 0.01, 1.0}{\strut vehicles}} {\setlength{\fboxsep}{0pt}\colorbox[Hsb]{202, 0.00, 1.0}{\strut .}} {\setlength{\fboxsep}{0pt}\colorbox[Hsb]{202, 0.00, 1.0}{\strut the}} {\setlength{\fboxsep}{0pt}\colorbox[Hsb]{202, 0.00, 1.0}{\strut two}} {\setlength{\fboxsep}{0pt}\colorbox[Hsb]{202, 0.00, 1.0}{\strut giant}} {\setlength{\fboxsep}{0pt}\colorbox[Hsb]{202, 0.04, 1.0}{\strut automakers}} {\setlength{\fboxsep}{0pt}\colorbox[Hsb]{202, 0.02, 1.0}{\strut say}} {\setlength{\fboxsep}{0pt}\colorbox[Hsb]{202, 0.00, 1.0}{\strut they}} {\setlength{\fboxsep}{0pt}\colorbox[Hsb]{202, 0.00, 1.0}{\strut have}} {\setlength{\fboxsep}{0pt}\colorbox[Hsb]{202, 0.00, 1.0}{\strut signed}} {\setlength{\fboxsep}{0pt}\colorbox[Hsb]{202, 0.00, 1.0}{\strut a}} {\setlength{\fboxsep}{0pt}\colorbox[Hsb]{202, 0.00, 1.0}{\strut memorandum}} {\setlength{\fboxsep}{0pt}\colorbox[Hsb]{202, 0.00, 1.0}{\strut of}} {\setlength{\fboxsep}{0pt}\colorbox[Hsb]{202, 0.00, 1.0}{\strut understanding}} 

\textbf{Adversarial}:{\setlength{\fboxsep}{0pt}\colorbox[Hsb]{202, 0.00, 1.0}{\strut general}} {\setlength{\fboxsep}{0pt}\colorbox[Hsb]{202, 0.00, 1.0}{\strut motors}} {\setlength{\fboxsep}{0pt}\colorbox[Hsb]{202, 0.00, 1.0}{\strut and}} {\setlength{\fboxsep}{0pt}\colorbox[Hsb]{202, 0.00, 1.0}{\strut daimlerchrysler}} {\setlength{\fboxsep}{0pt}\colorbox[Hsb]{202, 0.00, 1.0}{\strut say}} {\setlength{\fboxsep}{0pt}\colorbox[Hsb]{202, 0.00, 1.0}{\strut they}} {\setlength{\fboxsep}{0pt}\colorbox[Hsb]{202, 0.00, 1.0}{\strut \#}} {\setlength{\fboxsep}{0pt}\colorbox[Hsb]{202, 0.00, 1.0}{\strut qqq}} {\setlength{\fboxsep}{0pt}\colorbox[Hsb]{202, 0.00, 1.0}{\strut teaming}} {\setlength{\fboxsep}{0pt}\colorbox[Hsb]{202, 0.00, 1.0}{\strut up}} {\setlength{\fboxsep}{0pt}\colorbox[Hsb]{202, 0.00, 1.0}{\strut to}} {\setlength{\fboxsep}{0pt}\colorbox[Hsb]{202, 0.00, 1.0}{\strut develop}} {\setlength{\fboxsep}{0pt}\colorbox[Hsb]{202, 0.00, 1.0}{\strut hybrid}} {\setlength{\fboxsep}{0pt}\colorbox[Hsb]{202, 0.00, 1.0}{\strut technology}} {\setlength{\fboxsep}{0pt}\colorbox[Hsb]{202, 0.00, 1.0}{\strut for}} {\setlength{\fboxsep}{0pt}\colorbox[Hsb]{202, 0.00, 1.0}{\strut use}} {\setlength{\fboxsep}{0pt}\colorbox[Hsb]{202, 0.00, 1.0}{\strut in}} {\setlength{\fboxsep}{0pt}\colorbox[Hsb]{202, 0.50, 1.0}{\strut their}} {\setlength{\fboxsep}{0pt}\colorbox[Hsb]{202, 0.00, 1.0}{\strut vehicles}} {\setlength{\fboxsep}{0pt}\colorbox[Hsb]{202, 0.00, 1.0}{\strut .}} {\setlength{\fboxsep}{0pt}\colorbox[Hsb]{202, 0.00, 1.0}{\strut the}} {\setlength{\fboxsep}{0pt}\colorbox[Hsb]{202, 0.00, 1.0}{\strut two}} {\setlength{\fboxsep}{0pt}\colorbox[Hsb]{202, 0.00, 1.0}{\strut giant}} {\setlength{\fboxsep}{0pt}\colorbox[Hsb]{202, 0.00, 1.0}{\strut automakers}} {\setlength{\fboxsep}{0pt}\colorbox[Hsb]{202, 0.00, 1.0}{\strut say}} {\setlength{\fboxsep}{0pt}\colorbox[Hsb]{202, 0.00, 1.0}{\strut they}} {\setlength{\fboxsep}{0pt}\colorbox[Hsb]{202, 0.00, 1.0}{\strut have}} {\setlength{\fboxsep}{0pt}\colorbox[Hsb]{202, 0.00, 1.0}{\strut signed}} {\setlength{\fboxsep}{0pt}\colorbox[Hsb]{202, 0.00, 1.0}{\strut a}} {\setlength{\fboxsep}{0pt}\colorbox[Hsb]{202, 0.00, 1.0}{\strut memorandum}} {\setlength{\fboxsep}{0pt}\colorbox[Hsb]{202, 0.00, 1.0}{\strut of}} {\setlength{\fboxsep}{0pt}\colorbox[Hsb]{202, 0.00, 1.0}{\strut understanding}} {\setlength{\fboxsep}{0pt}\colorbox[Hsb]{202, 0.00, 1.0}{\strut .}}
\textbf{$\Delta\hat{y}$}: \emph{0.006}

\vspace{1em}
\par
\noindent\textbf{SNLI}

\textbf{Hypothesis}:a man is running on foot

\textbf{Original Premise Attention}:{\setlength{\fboxsep}{0pt}\colorbox[Hsb]{202, 0.00, 1.0}{\strut a}} {\setlength{\fboxsep}{0pt}\colorbox[Hsb]{202, 0.01, 1.0}{\strut man}} {\setlength{\fboxsep}{0pt}\colorbox[Hsb]{202, 0.00, 1.0}{\strut in}} {\setlength{\fboxsep}{0pt}\colorbox[Hsb]{202, 0.00, 1.0}{\strut a}} {\setlength{\fboxsep}{0pt}\colorbox[Hsb]{202, 0.00, 1.0}{\strut gray}} {\setlength{\fboxsep}{0pt}\colorbox[Hsb]{202, 0.00, 1.0}{\strut shirt}} {\setlength{\fboxsep}{0pt}\colorbox[Hsb]{202, 0.00, 1.0}{\strut and}} {\setlength{\fboxsep}{0pt}\colorbox[Hsb]{202, 0.00, 1.0}{\strut blue}} {\setlength{\fboxsep}{0pt}\colorbox[Hsb]{202, 0.01, 1.0}{\strut shorts}} {\setlength{\fboxsep}{0pt}\colorbox[Hsb]{202, 0.00, 1.0}{\strut is}} {\setlength{\fboxsep}{0pt}\colorbox[Hsb]{202, 0.17, 1.0}{\strut standing}} {\setlength{\fboxsep}{0pt}\colorbox[Hsb]{202, 0.01, 1.0}{\strut outside}} {\setlength{\fboxsep}{0pt}\colorbox[Hsb]{202, 0.00, 1.0}{\strut of}} {\setlength{\fboxsep}{0pt}\colorbox[Hsb]{202, 0.00, 1.0}{\strut an}} {\setlength{\fboxsep}{0pt}\colorbox[Hsb]{202, 0.00, 1.0}{\strut old}} {\setlength{\fboxsep}{0pt}\colorbox[Hsb]{202, 0.01, 1.0}{\strut fashioned}} {\setlength{\fboxsep}{0pt}\colorbox[Hsb]{202, 0.01, 1.0}{\strut ice}} {\setlength{\fboxsep}{0pt}\colorbox[Hsb]{202, 0.01, 1.0}{\strut cream}} {\setlength{\fboxsep}{0pt}\colorbox[Hsb]{202, 0.00, 1.0}{\strut shop}} {\setlength{\fboxsep}{0pt}\colorbox[Hsb]{202, 0.00, 1.0}{\strut named}} {\setlength{\fboxsep}{0pt}\colorbox[Hsb]{202, 0.00, 1.0}{\strut sara}} {\setlength{\fboxsep}{0pt}\colorbox[Hsb]{202, 0.00, 1.0}{\strut 's}} {\setlength{\fboxsep}{0pt}\colorbox[Hsb]{202, 0.00, 1.0}{\strut old}} {\setlength{\fboxsep}{0pt}\colorbox[Hsb]{202, 0.00, 1.0}{\strut fashioned}} {\setlength{\fboxsep}{0pt}\colorbox[Hsb]{202, 0.00, 1.0}{\strut ice}} {\setlength{\fboxsep}{0pt}\colorbox[Hsb]{202, 0.01, 1.0}{\strut cream}} {\setlength{\fboxsep}{0pt}\colorbox[Hsb]{202, 0.00, 1.0}{\strut ,}} {\setlength{\fboxsep}{0pt}\colorbox[Hsb]{202, 0.02, 1.0}{\strut holding}} {\setlength{\fboxsep}{0pt}\colorbox[Hsb]{202, 0.00, 1.0}{\strut his}} {\setlength{\fboxsep}{0pt}\colorbox[Hsb]{202, 0.13, 1.0}{\strut bike}} {\setlength{\fboxsep}{0pt}\colorbox[Hsb]{202, 0.00, 1.0}{\strut up}} {\setlength{\fboxsep}{0pt}\colorbox[Hsb]{202, 0.00, 1.0}{\strut ,}} {\setlength{\fboxsep}{0pt}\colorbox[Hsb]{202, 0.00, 1.0}{\strut with}} {\setlength{\fboxsep}{0pt}\colorbox[Hsb]{202, 0.00, 1.0}{\strut a}} {\setlength{\fboxsep}{0pt}\colorbox[Hsb]{202, 0.00, 1.0}{\strut wood}} {\setlength{\fboxsep}{0pt}\colorbox[Hsb]{202, 0.00, 1.0}{\strut like}} {\setlength{\fboxsep}{0pt}\colorbox[Hsb]{202, 0.00, 1.0}{\strut table}} {\setlength{\fboxsep}{0pt}\colorbox[Hsb]{202, 0.00, 1.0}{\strut ,}} {\setlength{\fboxsep}{0pt}\colorbox[Hsb]{202, 0.01, 1.0}{\strut chairs}} {\setlength{\fboxsep}{0pt}\colorbox[Hsb]{202, 0.00, 1.0}{\strut ,}} {\setlength{\fboxsep}{0pt}\colorbox[Hsb]{202, 0.00, 1.0}{\strut benches}} {\setlength{\fboxsep}{0pt}\colorbox[Hsb]{202, 0.00, 1.0}{\strut in}} {\setlength{\fboxsep}{0pt}\colorbox[Hsb]{202, 0.01, 1.0}{\strut front}} {\setlength{\fboxsep}{0pt}\colorbox[Hsb]{202, 0.00, 1.0}{\strut of}} {\setlength{\fboxsep}{0pt}\colorbox[Hsb]{202, 0.01, 1.0}{\strut him}} {\setlength{\fboxsep}{0pt}\colorbox[Hsb]{202, 0.01, 1.0}{\strut .}}

\textbf{Adversarial Premise Attention}:{\setlength{\fboxsep}{0pt}\colorbox[Hsb]{202, 0.00, 1.0}{\strut a}} {\setlength{\fboxsep}{0pt}\colorbox[Hsb]{202, 0.00, 1.0}{\strut man}} {\setlength{\fboxsep}{0pt}\colorbox[Hsb]{202, 0.00, 1.0}{\strut in}} {\setlength{\fboxsep}{0pt}\colorbox[Hsb]{202, 0.00, 1.0}{\strut a}} {\setlength{\fboxsep}{0pt}\colorbox[Hsb]{202, 0.00, 1.0}{\strut gray}} {\setlength{\fboxsep}{0pt}\colorbox[Hsb]{202, 0.50, 1.0}{\strut shirt}} {\setlength{\fboxsep}{0pt}\colorbox[Hsb]{202, 0.00, 1.0}{\strut and}} {\setlength{\fboxsep}{0pt}\colorbox[Hsb]{202, 0.00, 1.0}{\strut blue}} {\setlength{\fboxsep}{0pt}\colorbox[Hsb]{202, 0.00, 1.0}{\strut shorts}} {\setlength{\fboxsep}{0pt}\colorbox[Hsb]{202, 0.00, 1.0}{\strut is}} {\setlength{\fboxsep}{0pt}\colorbox[Hsb]{202, 0.00, 1.0}{\strut standing}} {\setlength{\fboxsep}{0pt}\colorbox[Hsb]{202, 0.00, 1.0}{\strut outside}} {\setlength{\fboxsep}{0pt}\colorbox[Hsb]{202, 0.00, 1.0}{\strut of}} {\setlength{\fboxsep}{0pt}\colorbox[Hsb]{202, 0.00, 1.0}{\strut an}} {\setlength{\fboxsep}{0pt}\colorbox[Hsb]{202, 0.00, 1.0}{\strut old}} {\setlength{\fboxsep}{0pt}\colorbox[Hsb]{202, 0.00, 1.0}{\strut fashioned}} {\setlength{\fboxsep}{0pt}\colorbox[Hsb]{202, 0.00, 1.0}{\strut ice}} {\setlength{\fboxsep}{0pt}\colorbox[Hsb]{202, 0.00, 1.0}{\strut cream}} {\setlength{\fboxsep}{0pt}\colorbox[Hsb]{202, 0.00, 1.0}{\strut shop}} {\setlength{\fboxsep}{0pt}\colorbox[Hsb]{202, 0.00, 1.0}{\strut named}} {\setlength{\fboxsep}{0pt}\colorbox[Hsb]{202, 0.00, 1.0}{\strut sara}} {\setlength{\fboxsep}{0pt}\colorbox[Hsb]{202, 0.00, 1.0}{\strut 's}} {\setlength{\fboxsep}{0pt}\colorbox[Hsb]{202, 0.00, 1.0}{\strut old}} {\setlength{\fboxsep}{0pt}\colorbox[Hsb]{202, 0.00, 1.0}{\strut fashioned}} {\setlength{\fboxsep}{0pt}\colorbox[Hsb]{202, 0.00, 1.0}{\strut ice}} {\setlength{\fboxsep}{0pt}\colorbox[Hsb]{202, 0.00, 1.0}{\strut cream}} {\setlength{\fboxsep}{0pt}\colorbox[Hsb]{202, 0.00, 1.0}{\strut ,}} {\setlength{\fboxsep}{0pt}\colorbox[Hsb]{202, 0.00, 1.0}{\strut holding}} {\setlength{\fboxsep}{0pt}\colorbox[Hsb]{202, 0.00, 1.0}{\strut his}} {\setlength{\fboxsep}{0pt}\colorbox[Hsb]{202, 0.00, 1.0}{\strut bike}} {\setlength{\fboxsep}{0pt}\colorbox[Hsb]{202, 0.00, 1.0}{\strut up}} {\setlength{\fboxsep}{0pt}\colorbox[Hsb]{202, 0.00, 1.0}{\strut ,}} {\setlength{\fboxsep}{0pt}\colorbox[Hsb]{202, 0.00, 1.0}{\strut with}} {\setlength{\fboxsep}{0pt}\colorbox[Hsb]{202, 0.00, 1.0}{\strut a}} {\setlength{\fboxsep}{0pt}\colorbox[Hsb]{202, 0.00, 1.0}{\strut wood}} {\setlength{\fboxsep}{0pt}\colorbox[Hsb]{202, 0.00, 1.0}{\strut like}} {\setlength{\fboxsep}{0pt}\colorbox[Hsb]{202, 0.00, 1.0}{\strut table}} {\setlength{\fboxsep}{0pt}\colorbox[Hsb]{202, 0.00, 1.0}{\strut ,}} {\setlength{\fboxsep}{0pt}\colorbox[Hsb]{202, 0.00, 1.0}{\strut chairs}} {\setlength{\fboxsep}{0pt}\colorbox[Hsb]{202, 0.00, 1.0}{\strut ,}} {\setlength{\fboxsep}{0pt}\colorbox[Hsb]{202, 0.00, 1.0}{\strut benches}} {\setlength{\fboxsep}{0pt}\colorbox[Hsb]{202, 0.00, 1.0}{\strut in}} {\setlength{\fboxsep}{0pt}\colorbox[Hsb]{202, 0.00, 1.0}{\strut front}} {\setlength{\fboxsep}{0pt}\colorbox[Hsb]{202, 0.00, 1.0}{\strut of}} {\setlength{\fboxsep}{0pt}\colorbox[Hsb]{202, 0.00, 1.0}{\strut him}} {\setlength{\fboxsep}{0pt}\colorbox[Hsb]{202, 0.00, 1.0}{\strut .}}
\textbf{$\Delta\hat{y}$}: \emph{0.002}

\vspace{1em}
\par
\noindent\textbf{Babi Task 1}

\textbf{Question}: Where is Sandra ?

\textbf{Original Attention}:{\setlength{\fboxsep}{0pt}\colorbox[Hsb]{202, 0.00, 1.0}{\strut John}} {\setlength{\fboxsep}{0pt}\colorbox[Hsb]{202, 0.00, 1.0}{\strut travelled}} {\setlength{\fboxsep}{0pt}\colorbox[Hsb]{202, 0.00, 1.0}{\strut to}} {\setlength{\fboxsep}{0pt}\colorbox[Hsb]{202, 0.00, 1.0}{\strut the}} {\setlength{\fboxsep}{0pt}\colorbox[Hsb]{202, 0.00, 1.0}{\strut garden}} {\setlength{\fboxsep}{0pt}\colorbox[Hsb]{202, 0.00, 1.0}{\strut .}} {\setlength{\fboxsep}{0pt}\colorbox[Hsb]{202, 0.00, 1.0}{\strut Sandra}} {\setlength{\fboxsep}{0pt}\colorbox[Hsb]{202, 0.00, 1.0}{\strut travelled}} {\setlength{\fboxsep}{0pt}\colorbox[Hsb]{202, 0.00, 1.0}{\strut to}} {\setlength{\fboxsep}{0pt}\colorbox[Hsb]{202, 0.00, 1.0}{\strut the}} {\setlength{\fboxsep}{0pt}\colorbox[Hsb]{202, 0.50, 1.0}{\strut garden}}

\textbf{Adversarial Attention}:{\setlength{\fboxsep}{0pt}\colorbox[Hsb]{202, 0.00, 1.0}{\strut John}} {\setlength{\fboxsep}{0pt}\colorbox[Hsb]{202, 0.00, 1.0}{\strut travelled}} {\setlength{\fboxsep}{0pt}\colorbox[Hsb]{202, 0.01, 1.0}{\strut to}} {\setlength{\fboxsep}{0pt}\colorbox[Hsb]{202, 0.01, 1.0}{\strut the}} {\setlength{\fboxsep}{0pt}\colorbox[Hsb]{202, 0.30, 1.0}{\strut garden}} {\setlength{\fboxsep}{0pt}\colorbox[Hsb]{202, 0.17, 1.0}{\strut .}} {\setlength{\fboxsep}{0pt}\colorbox[Hsb]{202, 0.00, 1.0}{\strut Sandra}} {\setlength{\fboxsep}{0pt}\colorbox[Hsb]{202, 0.00, 1.0}{\strut travelled}} {\setlength{\fboxsep}{0pt}\colorbox[Hsb]{202, 0.00, 1.0}{\strut to}} {\setlength{\fboxsep}{0pt}\colorbox[Hsb]{202, 0.00, 1.0}{\strut the}} {\setlength{\fboxsep}{0pt}\colorbox[Hsb]{202, 0.00, 1.0}{\strut garden}}
\textbf{$\Delta\hat{y}$}: \emph{0.003}

\vspace{1em}
\par
\noindent\textbf{CNN-QA}

\textbf{Question}:federal education minister @placeholder visited a @entity15 store in @entity17 , saw cameras

\textbf{Original}:{\setlength{\fboxsep}{0pt}\colorbox[Hsb]{202, 0.00, 1.0}{\strut @entity1}} {\setlength{\fboxsep}{0pt}\colorbox[Hsb]{202, 0.00, 1.0}{\strut ,}} {\setlength{\fboxsep}{0pt}\colorbox[Hsb]{202, 0.00, 1.0}{\strut @entity2}} {\setlength{\fboxsep}{0pt}\colorbox[Hsb]{202, 0.00, 1.0}{\strut (}} {\setlength{\fboxsep}{0pt}\colorbox[Hsb]{202, 0.00, 1.0}{\strut @entity3}} {\setlength{\fboxsep}{0pt}\colorbox[Hsb]{202, 0.00, 1.0}{\strut )}} {\setlength{\fboxsep}{0pt}\colorbox[Hsb]{202, 0.00, 1.0}{\strut police}} {\setlength{\fboxsep}{0pt}\colorbox[Hsb]{202, 0.00, 1.0}{\strut have}} {\setlength{\fboxsep}{0pt}\colorbox[Hsb]{202, 0.00, 1.0}{\strut arrested}} {\setlength{\fboxsep}{0pt}\colorbox[Hsb]{202, 0.00, 1.0}{\strut four}} {\setlength{\fboxsep}{0pt}\colorbox[Hsb]{202, 0.00, 1.0}{\strut employees}} {\setlength{\fboxsep}{0pt}\colorbox[Hsb]{202, 0.00, 1.0}{\strut of}} {\setlength{\fboxsep}{0pt}\colorbox[Hsb]{202, 0.00, 1.0}{\strut a}} {\setlength{\fboxsep}{0pt}\colorbox[Hsb]{202, 0.00, 1.0}{\strut popular}} {\setlength{\fboxsep}{0pt}\colorbox[Hsb]{202, 0.00, 1.0}{\strut @entity2}} {\setlength{\fboxsep}{0pt}\colorbox[Hsb]{202, 0.00, 1.0}{\strut ethnic}} {\setlength{\fboxsep}{0pt}\colorbox[Hsb]{202, 0.00, 1.0}{\strut -}} {\setlength{\fboxsep}{0pt}\colorbox[Hsb]{202, 0.00, 1.0}{\strut wear}} {\setlength{\fboxsep}{0pt}\colorbox[Hsb]{202, 0.00, 1.0}{\strut chain}} {\setlength{\fboxsep}{0pt}\colorbox[Hsb]{202, 0.00, 1.0}{\strut after}} {\setlength{\fboxsep}{0pt}\colorbox[Hsb]{202, 0.00, 1.0}{\strut a}} {\setlength{\fboxsep}{0pt}\colorbox[Hsb]{202, 0.00, 1.0}{\strut minister}} {\setlength{\fboxsep}{0pt}\colorbox[Hsb]{202, 0.01, 1.0}{\strut spotted}} {\setlength{\fboxsep}{0pt}\colorbox[Hsb]{202, 0.00, 1.0}{\strut a}} {\setlength{\fboxsep}{0pt}\colorbox[Hsb]{202, 0.00, 1.0}{\strut security}} {\setlength{\fboxsep}{0pt}\colorbox[Hsb]{202, 0.00, 1.0}{\strut camera}} {\setlength{\fboxsep}{0pt}\colorbox[Hsb]{202, 0.00, 1.0}{\strut overlooking}} {\setlength{\fboxsep}{0pt}\colorbox[Hsb]{202, 0.00, 1.0}{\strut the}} {\setlength{\fboxsep}{0pt}\colorbox[Hsb]{202, 0.00, 1.0}{\strut changing}} {\setlength{\fboxsep}{0pt}\colorbox[Hsb]{202, 0.00, 1.0}{\strut room}} {\setlength{\fboxsep}{0pt}\colorbox[Hsb]{202, 0.00, 1.0}{\strut of}} {\setlength{\fboxsep}{0pt}\colorbox[Hsb]{202, 0.00, 1.0}{\strut one}} {\setlength{\fboxsep}{0pt}\colorbox[Hsb]{202, 0.00, 1.0}{\strut of}} {\setlength{\fboxsep}{0pt}\colorbox[Hsb]{202, 0.00, 1.0}{\strut its}} {\setlength{\fboxsep}{0pt}\colorbox[Hsb]{202, 0.00, 1.0}{\strut stores}} {\setlength{\fboxsep}{0pt}\colorbox[Hsb]{202, 0.00, 1.0}{\strut .}} {\setlength{\fboxsep}{0pt}\colorbox[Hsb]{202, 0.00, 1.0}{\strut federal}} {\setlength{\fboxsep}{0pt}\colorbox[Hsb]{202, 0.00, 1.0}{\strut education}} {\setlength{\fboxsep}{0pt}\colorbox[Hsb]{202, 0.00, 1.0}{\strut minister}} {\setlength{\fboxsep}{0pt}\colorbox[Hsb]{202, 0.43, 1.0}{\strut @entity13}} {\setlength{\fboxsep}{0pt}\colorbox[Hsb]{202, 0.00, 1.0}{\strut was}} {\setlength{\fboxsep}{0pt}\colorbox[Hsb]{202, 0.00, 1.0}{\strut visiting}} {\setlength{\fboxsep}{0pt}\colorbox[Hsb]{202, 0.00, 1.0}{\strut a}} {\setlength{\fboxsep}{0pt}\colorbox[Hsb]{202, 0.00, 1.0}{\strut @entity15}} {\setlength{\fboxsep}{0pt}\colorbox[Hsb]{202, 0.00, 1.0}{\strut outlet}} {\setlength{\fboxsep}{0pt}\colorbox[Hsb]{202, 0.00, 1.0}{\strut in}} {\setlength{\fboxsep}{0pt}\colorbox[Hsb]{202, 0.00, 1.0}{\strut the}} {\setlength{\fboxsep}{0pt}\colorbox[Hsb]{202, 0.00, 1.0}{\strut tourist}} {\setlength{\fboxsep}{0pt}\colorbox[Hsb]{202, 0.00, 1.0}{\strut resort}} {\setlength{\fboxsep}{0pt}\colorbox[Hsb]{202, 0.00, 1.0}{\strut state}} {\setlength{\fboxsep}{0pt}\colorbox[Hsb]{202, 0.00, 1.0}{\strut of}} {\setlength{\fboxsep}{0pt}\colorbox[Hsb]{202, 0.00, 1.0}{\strut @entity17}} {\setlength{\fboxsep}{0pt}\colorbox[Hsb]{202, 0.00, 1.0}{\strut on}} {\setlength{\fboxsep}{0pt}\colorbox[Hsb]{202, 0.00, 1.0}{\strut friday}} {\setlength{\fboxsep}{0pt}\colorbox[Hsb]{202, 0.00, 1.0}{\strut when}} {\setlength{\fboxsep}{0pt}\colorbox[Hsb]{202, 0.00, 1.0}{\strut she}} {\setlength{\fboxsep}{0pt}\colorbox[Hsb]{202, 0.00, 1.0}{\strut discovered}} {\setlength{\fboxsep}{0pt}\colorbox[Hsb]{202, 0.00, 1.0}{\strut a}} {\setlength{\fboxsep}{0pt}\colorbox[Hsb]{202, 0.00, 1.0}{\strut surveillance}} {\setlength{\fboxsep}{0pt}\colorbox[Hsb]{202, 0.00, 1.0}{\strut camera}} {\setlength{\fboxsep}{0pt}\colorbox[Hsb]{202, 0.00, 1.0}{\strut pointed}} {\setlength{\fboxsep}{0pt}\colorbox[Hsb]{202, 0.00, 1.0}{\strut at}} {\setlength{\fboxsep}{0pt}\colorbox[Hsb]{202, 0.00, 1.0}{\strut the}} {\setlength{\fboxsep}{0pt}\colorbox[Hsb]{202, 0.00, 1.0}{\strut changing}} {\setlength{\fboxsep}{0pt}\colorbox[Hsb]{202, 0.00, 1.0}{\strut room}} {\setlength{\fboxsep}{0pt}\colorbox[Hsb]{202, 0.00, 1.0}{\strut ,}} {\setlength{\fboxsep}{0pt}\colorbox[Hsb]{202, 0.00, 1.0}{\strut police}} {\setlength{\fboxsep}{0pt}\colorbox[Hsb]{202, 0.00, 1.0}{\strut said}} {\setlength{\fboxsep}{0pt}\colorbox[Hsb]{202, 0.00, 1.0}{\strut .}} {\setlength{\fboxsep}{0pt}\colorbox[Hsb]{202, 0.00, 1.0}{\strut four}} {\setlength{\fboxsep}{0pt}\colorbox[Hsb]{202, 0.00, 1.0}{\strut employees}} {\setlength{\fboxsep}{0pt}\colorbox[Hsb]{202, 0.00, 1.0}{\strut of}} {\setlength{\fboxsep}{0pt}\colorbox[Hsb]{202, 0.00, 1.0}{\strut the}} {\setlength{\fboxsep}{0pt}\colorbox[Hsb]{202, 0.00, 1.0}{\strut store}} {\setlength{\fboxsep}{0pt}\colorbox[Hsb]{202, 0.00, 1.0}{\strut have}} {\setlength{\fboxsep}{0pt}\colorbox[Hsb]{202, 0.00, 1.0}{\strut been}} {\setlength{\fboxsep}{0pt}\colorbox[Hsb]{202, 0.00, 1.0}{\strut arrested}} {\setlength{\fboxsep}{0pt}\colorbox[Hsb]{202, 0.00, 1.0}{\strut ,}} {\setlength{\fboxsep}{0pt}\colorbox[Hsb]{202, 0.00, 1.0}{\strut but}} {\setlength{\fboxsep}{0pt}\colorbox[Hsb]{202, 0.00, 1.0}{\strut its}} {\setlength{\fboxsep}{0pt}\colorbox[Hsb]{202, 0.00, 1.0}{\strut manager}} {\setlength{\fboxsep}{0pt}\colorbox[Hsb]{202, 0.00, 1.0}{\strut --}} {\setlength{\fboxsep}{0pt}\colorbox[Hsb]{202, 0.00, 1.0}{\strut herself}} {\setlength{\fboxsep}{0pt}\colorbox[Hsb]{202, 0.00, 1.0}{\strut a}} {\setlength{\fboxsep}{0pt}\colorbox[Hsb]{202, 0.00, 1.0}{\strut woman}} {\setlength{\fboxsep}{0pt}\colorbox[Hsb]{202, 0.00, 1.0}{\strut --}} {\setlength{\fboxsep}{0pt}\colorbox[Hsb]{202, 0.00, 1.0}{\strut was}} {\setlength{\fboxsep}{0pt}\colorbox[Hsb]{202, 0.00, 1.0}{\strut still}} {\setlength{\fboxsep}{0pt}\colorbox[Hsb]{202, 0.00, 1.0}{\strut at}} {\setlength{\fboxsep}{0pt}\colorbox[Hsb]{202, 0.00, 1.0}{\strut large}} {\setlength{\fboxsep}{0pt}\colorbox[Hsb]{202, 0.00, 1.0}{\strut saturday}} {\setlength{\fboxsep}{0pt}\colorbox[Hsb]{202, 0.00, 1.0}{\strut ,}} {\setlength{\fboxsep}{0pt}\colorbox[Hsb]{202, 0.00, 1.0}{\strut said}} {\setlength{\fboxsep}{0pt}\colorbox[Hsb]{202, 0.00, 1.0}{\strut @entity17}} {\setlength{\fboxsep}{0pt}\colorbox[Hsb]{202, 0.00, 1.0}{\strut police}} {\setlength{\fboxsep}{0pt}\colorbox[Hsb]{202, 0.00, 1.0}{\strut superintendent}} {\setlength{\fboxsep}{0pt}\colorbox[Hsb]{202, 0.03, 1.0}{\strut @entity25}} {\setlength{\fboxsep}{0pt}\colorbox[Hsb]{202, 0.00, 1.0}{\strut .}} {\setlength{\fboxsep}{0pt}\colorbox[Hsb]{202, 0.00, 1.0}{\strut state}} {\setlength{\fboxsep}{0pt}\colorbox[Hsb]{202, 0.00, 1.0}{\strut authorities}} {\setlength{\fboxsep}{0pt}\colorbox[Hsb]{202, 0.00, 1.0}{\strut launched}} {\setlength{\fboxsep}{0pt}\colorbox[Hsb]{202, 0.00, 1.0}{\strut their}} {\setlength{\fboxsep}{0pt}\colorbox[Hsb]{202, 0.00, 1.0}{\strut investigation}} {\setlength{\fboxsep}{0pt}\colorbox[Hsb]{202, 0.00, 1.0}{\strut right}} {\setlength{\fboxsep}{0pt}\colorbox[Hsb]{202, 0.00, 1.0}{\strut after}} {\setlength{\fboxsep}{0pt}\colorbox[Hsb]{202, 0.00, 1.0}{\strut @entity13}} {\setlength{\fboxsep}{0pt}\colorbox[Hsb]{202, 0.00, 1.0}{\strut levied}} {\setlength{\fboxsep}{0pt}\colorbox[Hsb]{202, 0.00, 1.0}{\strut her}} {\setlength{\fboxsep}{0pt}\colorbox[Hsb]{202, 0.00, 1.0}{\strut accusation}} {\setlength{\fboxsep}{0pt}\colorbox[Hsb]{202, 0.00, 1.0}{\strut .}} {\setlength{\fboxsep}{0pt}\colorbox[Hsb]{202, 0.00, 1.0}{\strut they}} {\setlength{\fboxsep}{0pt}\colorbox[Hsb]{202, 0.00, 1.0}{\strut found}} {\setlength{\fboxsep}{0pt}\colorbox[Hsb]{202, 0.00, 1.0}{\strut an}} {\setlength{\fboxsep}{0pt}\colorbox[Hsb]{202, 0.00, 1.0}{\strut overhead}} {\setlength{\fboxsep}{0pt}\colorbox[Hsb]{202, 0.00, 1.0}{\strut camera}} {\setlength{\fboxsep}{0pt}\colorbox[Hsb]{202, 0.00, 1.0}{\strut that}} {\setlength{\fboxsep}{0pt}\colorbox[Hsb]{202, 0.00, 1.0}{\strut the}} {\setlength{\fboxsep}{0pt}\colorbox[Hsb]{202, 0.00, 1.0}{\strut minister}} {\setlength{\fboxsep}{0pt}\colorbox[Hsb]{202, 0.00, 1.0}{\strut had}} {\setlength{\fboxsep}{0pt}\colorbox[Hsb]{202, 0.00, 1.0}{\strut spotted}} {\setlength{\fboxsep}{0pt}\colorbox[Hsb]{202, 0.00, 1.0}{\strut and}} {\setlength{\fboxsep}{0pt}\colorbox[Hsb]{202, 0.00, 1.0}{\strut determined}} {\setlength{\fboxsep}{0pt}\colorbox[Hsb]{202, 0.00, 1.0}{\strut that}} {\setlength{\fboxsep}{0pt}\colorbox[Hsb]{202, 0.00, 1.0}{\strut it}} {\setlength{\fboxsep}{0pt}\colorbox[Hsb]{202, 0.00, 1.0}{\strut was}} {\setlength{\fboxsep}{0pt}\colorbox[Hsb]{202, 0.00, 1.0}{\strut indeed}} {\setlength{\fboxsep}{0pt}\colorbox[Hsb]{202, 0.00, 1.0}{\strut able}} {\setlength{\fboxsep}{0pt}\colorbox[Hsb]{202, 0.00, 1.0}{\strut to}} {\setlength{\fboxsep}{0pt}\colorbox[Hsb]{202, 0.00, 1.0}{\strut take}} {\setlength{\fboxsep}{0pt}\colorbox[Hsb]{202, 0.00, 1.0}{\strut photos}} {\setlength{\fboxsep}{0pt}\colorbox[Hsb]{202, 0.00, 1.0}{\strut of}} {\setlength{\fboxsep}{0pt}\colorbox[Hsb]{202, 0.00, 1.0}{\strut customers}} {\setlength{\fboxsep}{0pt}\colorbox[Hsb]{202, 0.00, 1.0}{\strut using}} {\setlength{\fboxsep}{0pt}\colorbox[Hsb]{202, 0.00, 1.0}{\strut the}} {\setlength{\fboxsep}{0pt}\colorbox[Hsb]{202, 0.00, 1.0}{\strut store}} {\setlength{\fboxsep}{0pt}\colorbox[Hsb]{202, 0.00, 1.0}{\strut 's}} {\setlength{\fboxsep}{0pt}\colorbox[Hsb]{202, 0.00, 1.0}{\strut changing}} {\setlength{\fboxsep}{0pt}\colorbox[Hsb]{202, 0.00, 1.0}{\strut room}} {\setlength{\fboxsep}{0pt}\colorbox[Hsb]{202, 0.00, 1.0}{\strut ,}} {\setlength{\fboxsep}{0pt}\colorbox[Hsb]{202, 0.00, 1.0}{\strut according}} {\setlength{\fboxsep}{0pt}\colorbox[Hsb]{202, 0.00, 1.0}{\strut to}} {\setlength{\fboxsep}{0pt}\colorbox[Hsb]{202, 0.00, 1.0}{\strut @entity25}} {\setlength{\fboxsep}{0pt}\colorbox[Hsb]{202, 0.00, 1.0}{\strut .}} {\setlength{\fboxsep}{0pt}\colorbox[Hsb]{202, 0.00, 1.0}{\strut after}} {\setlength{\fboxsep}{0pt}\colorbox[Hsb]{202, 0.00, 1.0}{\strut the}} {\setlength{\fboxsep}{0pt}\colorbox[Hsb]{202, 0.00, 1.0}{\strut incident}} {\setlength{\fboxsep}{0pt}\colorbox[Hsb]{202, 0.00, 1.0}{\strut ,}} {\setlength{\fboxsep}{0pt}\colorbox[Hsb]{202, 0.00, 1.0}{\strut authorities}} {\setlength{\fboxsep}{0pt}\colorbox[Hsb]{202, 0.00, 1.0}{\strut sealed}} {\setlength{\fboxsep}{0pt}\colorbox[Hsb]{202, 0.00, 1.0}{\strut off}} {\setlength{\fboxsep}{0pt}\colorbox[Hsb]{202, 0.00, 1.0}{\strut the}} {\setlength{\fboxsep}{0pt}\colorbox[Hsb]{202, 0.00, 1.0}{\strut store}} {\setlength{\fboxsep}{0pt}\colorbox[Hsb]{202, 0.00, 1.0}{\strut and}} {\setlength{\fboxsep}{0pt}\colorbox[Hsb]{202, 0.00, 1.0}{\strut summoned}} {\setlength{\fboxsep}{0pt}\colorbox[Hsb]{202, 0.00, 1.0}{\strut six}} {\setlength{\fboxsep}{0pt}\colorbox[Hsb]{202, 0.00, 1.0}{\strut top}} {\setlength{\fboxsep}{0pt}\colorbox[Hsb]{202, 0.00, 1.0}{\strut officials}} {\setlength{\fboxsep}{0pt}\colorbox[Hsb]{202, 0.00, 1.0}{\strut from}} {\setlength{\fboxsep}{0pt}\colorbox[Hsb]{202, 0.00, 1.0}{\strut @entity15}} {\setlength{\fboxsep}{0pt}\colorbox[Hsb]{202, 0.00, 1.0}{\strut ,}} {\setlength{\fboxsep}{0pt}\colorbox[Hsb]{202, 0.00, 1.0}{\strut he}} {\setlength{\fboxsep}{0pt}\colorbox[Hsb]{202, 0.00, 1.0}{\strut said}} {\setlength{\fboxsep}{0pt}\colorbox[Hsb]{202, 0.00, 1.0}{\strut .}} {\setlength{\fboxsep}{0pt}\colorbox[Hsb]{202, 0.00, 1.0}{\strut the}} {\setlength{\fboxsep}{0pt}\colorbox[Hsb]{202, 0.00, 1.0}{\strut arrested}} {\setlength{\fboxsep}{0pt}\colorbox[Hsb]{202, 0.00, 1.0}{\strut staff}} {\setlength{\fboxsep}{0pt}\colorbox[Hsb]{202, 0.00, 1.0}{\strut have}} {\setlength{\fboxsep}{0pt}\colorbox[Hsb]{202, 0.00, 1.0}{\strut been}} {\setlength{\fboxsep}{0pt}\colorbox[Hsb]{202, 0.00, 1.0}{\strut charged}} {\setlength{\fboxsep}{0pt}\colorbox[Hsb]{202, 0.00, 1.0}{\strut with}} {\setlength{\fboxsep}{0pt}\colorbox[Hsb]{202, 0.00, 1.0}{\strut voyeurism}} {\setlength{\fboxsep}{0pt}\colorbox[Hsb]{202, 0.00, 1.0}{\strut and}} {\setlength{\fboxsep}{0pt}\colorbox[Hsb]{202, 0.00, 1.0}{\strut breach}} {\setlength{\fboxsep}{0pt}\colorbox[Hsb]{202, 0.00, 1.0}{\strut of}} {\setlength{\fboxsep}{0pt}\colorbox[Hsb]{202, 0.00, 1.0}{\strut privacy}} {\setlength{\fboxsep}{0pt}\colorbox[Hsb]{202, 0.00, 1.0}{\strut ,}} {\setlength{\fboxsep}{0pt}\colorbox[Hsb]{202, 0.00, 1.0}{\strut according}} {\setlength{\fboxsep}{0pt}\colorbox[Hsb]{202, 0.00, 1.0}{\strut to}} {\setlength{\fboxsep}{0pt}\colorbox[Hsb]{202, 0.00, 1.0}{\strut the}} {\setlength{\fboxsep}{0pt}\colorbox[Hsb]{202, 0.00, 1.0}{\strut police}} {\setlength{\fboxsep}{0pt}\colorbox[Hsb]{202, 0.00, 1.0}{\strut .}} {\setlength{\fboxsep}{0pt}\colorbox[Hsb]{202, 0.00, 1.0}{\strut if}} {\setlength{\fboxsep}{0pt}\colorbox[Hsb]{202, 0.00, 1.0}{\strut convicted}} {\setlength{\fboxsep}{0pt}\colorbox[Hsb]{202, 0.00, 1.0}{\strut ,}} {\setlength{\fboxsep}{0pt}\colorbox[Hsb]{202, 0.00, 1.0}{\strut they}} {\setlength{\fboxsep}{0pt}\colorbox[Hsb]{202, 0.00, 1.0}{\strut could}} {\setlength{\fboxsep}{0pt}\colorbox[Hsb]{202, 0.00, 1.0}{\strut spend}} {\setlength{\fboxsep}{0pt}\colorbox[Hsb]{202, 0.00, 1.0}{\strut up}} {\setlength{\fboxsep}{0pt}\colorbox[Hsb]{202, 0.00, 1.0}{\strut to}} {\setlength{\fboxsep}{0pt}\colorbox[Hsb]{202, 0.00, 1.0}{\strut three}} {\setlength{\fboxsep}{0pt}\colorbox[Hsb]{202, 0.00, 1.0}{\strut years}} {\setlength{\fboxsep}{0pt}\colorbox[Hsb]{202, 0.00, 1.0}{\strut in}} {\setlength{\fboxsep}{0pt}\colorbox[Hsb]{202, 0.00, 1.0}{\strut jail}} {\setlength{\fboxsep}{0pt}\colorbox[Hsb]{202, 0.00, 1.0}{\strut ,}} {\setlength{\fboxsep}{0pt}\colorbox[Hsb]{202, 0.00, 1.0}{\strut @entity25}} {\setlength{\fboxsep}{0pt}\colorbox[Hsb]{202, 0.00, 1.0}{\strut said}} {\setlength{\fboxsep}{0pt}\colorbox[Hsb]{202, 0.00, 1.0}{\strut .}} {\setlength{\fboxsep}{0pt}\colorbox[Hsb]{202, 0.00, 1.0}{\strut officials}} {\setlength{\fboxsep}{0pt}\colorbox[Hsb]{202, 0.00, 1.0}{\strut from}} {\setlength{\fboxsep}{0pt}\colorbox[Hsb]{202, 0.00, 1.0}{\strut @entity15}} {\setlength{\fboxsep}{0pt}\colorbox[Hsb]{202, 0.00, 1.0}{\strut --}} {\setlength{\fboxsep}{0pt}\colorbox[Hsb]{202, 0.00, 1.0}{\strut which}} {\setlength{\fboxsep}{0pt}\colorbox[Hsb]{202, 0.00, 1.0}{\strut sells}} {\setlength{\fboxsep}{0pt}\colorbox[Hsb]{202, 0.00, 1.0}{\strut ethnic}} {\setlength{\fboxsep}{0pt}\colorbox[Hsb]{202, 0.00, 1.0}{\strut garments}} {\setlength{\fboxsep}{0pt}\colorbox[Hsb]{202, 0.00, 1.0}{\strut ,}} {\setlength{\fboxsep}{0pt}\colorbox[Hsb]{202, 0.00, 1.0}{\strut fabrics}} {\setlength{\fboxsep}{0pt}\colorbox[Hsb]{202, 0.00, 1.0}{\strut and}} {\setlength{\fboxsep}{0pt}\colorbox[Hsb]{202, 0.00, 1.0}{\strut other}} {\setlength{\fboxsep}{0pt}\colorbox[Hsb]{202, 0.00, 1.0}{\strut products}} {\setlength{\fboxsep}{0pt}\colorbox[Hsb]{202, 0.00, 1.0}{\strut --}} {\setlength{\fboxsep}{0pt}\colorbox[Hsb]{202, 0.00, 1.0}{\strut are}} {\setlength{\fboxsep}{0pt}\colorbox[Hsb]{202, 0.00, 1.0}{\strut heading}} {\setlength{\fboxsep}{0pt}\colorbox[Hsb]{202, 0.00, 1.0}{\strut to}} {\setlength{\fboxsep}{0pt}\colorbox[Hsb]{202, 0.00, 1.0}{\strut @entity17}} {\setlength{\fboxsep}{0pt}\colorbox[Hsb]{202, 0.00, 1.0}{\strut to}} {\setlength{\fboxsep}{0pt}\colorbox[Hsb]{202, 0.00, 1.0}{\strut work}} {\setlength{\fboxsep}{0pt}\colorbox[Hsb]{202, 0.00, 1.0}{\strut with}} {\setlength{\fboxsep}{0pt}\colorbox[Hsb]{202, 0.00, 1.0}{\strut investigators}} {\setlength{\fboxsep}{0pt}\colorbox[Hsb]{202, 0.00, 1.0}{\strut ,}} {\setlength{\fboxsep}{0pt}\colorbox[Hsb]{202, 0.00, 1.0}{\strut according}} {\setlength{\fboxsep}{0pt}\colorbox[Hsb]{202, 0.00, 1.0}{\strut to}} {\setlength{\fboxsep}{0pt}\colorbox[Hsb]{202, 0.00, 1.0}{\strut the}} {\setlength{\fboxsep}{0pt}\colorbox[Hsb]{202, 0.00, 1.0}{\strut company}} {\setlength{\fboxsep}{0pt}\colorbox[Hsb]{202, 0.00, 1.0}{\strut .}} {\setlength{\fboxsep}{0pt}\colorbox[Hsb]{202, 0.00, 1.0}{\strut "}} {\setlength{\fboxsep}{0pt}\colorbox[Hsb]{202, 0.01, 1.0}{\strut @entity15}} {\setlength{\fboxsep}{0pt}\colorbox[Hsb]{202, 0.00, 1.0}{\strut is}} {\setlength{\fboxsep}{0pt}\colorbox[Hsb]{202, 0.00, 1.0}{\strut deeply}} {\setlength{\fboxsep}{0pt}\colorbox[Hsb]{202, 0.00, 1.0}{\strut concerned}} {\setlength{\fboxsep}{0pt}\colorbox[Hsb]{202, 0.00, 1.0}{\strut and}} {\setlength{\fboxsep}{0pt}\colorbox[Hsb]{202, 0.00, 1.0}{\strut shocked}} {\setlength{\fboxsep}{0pt}\colorbox[Hsb]{202, 0.00, 1.0}{\strut at}} {\setlength{\fboxsep}{0pt}\colorbox[Hsb]{202, 0.00, 1.0}{\strut this}} {\setlength{\fboxsep}{0pt}\colorbox[Hsb]{202, 0.00, 1.0}{\strut allegation}} {\setlength{\fboxsep}{0pt}\colorbox[Hsb]{202, 0.00, 1.0}{\strut ,}} {\setlength{\fboxsep}{0pt}\colorbox[Hsb]{202, 0.00, 1.0}{\strut "}} {\setlength{\fboxsep}{0pt}\colorbox[Hsb]{202, 0.00, 1.0}{\strut the}} {\setlength{\fboxsep}{0pt}\colorbox[Hsb]{202, 0.00, 1.0}{\strut company}} {\setlength{\fboxsep}{0pt}\colorbox[Hsb]{202, 0.00, 1.0}{\strut said}} {\setlength{\fboxsep}{0pt}\colorbox[Hsb]{202, 0.00, 1.0}{\strut in}} {\setlength{\fboxsep}{0pt}\colorbox[Hsb]{202, 0.00, 1.0}{\strut a}} {\setlength{\fboxsep}{0pt}\colorbox[Hsb]{202, 0.00, 1.0}{\strut statement}} {\setlength{\fboxsep}{0pt}\colorbox[Hsb]{202, 0.00, 1.0}{\strut .}} {\setlength{\fboxsep}{0pt}\colorbox[Hsb]{202, 0.00, 1.0}{\strut "}} {\setlength{\fboxsep}{0pt}\colorbox[Hsb]{202, 0.00, 1.0}{\strut we}} {\setlength{\fboxsep}{0pt}\colorbox[Hsb]{202, 0.00, 1.0}{\strut are}} {\setlength{\fboxsep}{0pt}\colorbox[Hsb]{202, 0.00, 1.0}{\strut in}} {\setlength{\fboxsep}{0pt}\colorbox[Hsb]{202, 0.00, 1.0}{\strut the}} {\setlength{\fboxsep}{0pt}\colorbox[Hsb]{202, 0.00, 1.0}{\strut process}} {\setlength{\fboxsep}{0pt}\colorbox[Hsb]{202, 0.00, 1.0}{\strut of}} {\setlength{\fboxsep}{0pt}\colorbox[Hsb]{202, 0.00, 1.0}{\strut investigating}} {\setlength{\fboxsep}{0pt}\colorbox[Hsb]{202, 0.00, 1.0}{\strut this}} {\setlength{\fboxsep}{0pt}\colorbox[Hsb]{202, 0.00, 1.0}{\strut internally}} {\setlength{\fboxsep}{0pt}\colorbox[Hsb]{202, 0.00, 1.0}{\strut and}} {\setlength{\fboxsep}{0pt}\colorbox[Hsb]{202, 0.00, 1.0}{\strut will}} {\setlength{\fboxsep}{0pt}\colorbox[Hsb]{202, 0.00, 1.0}{\strut be}} {\setlength{\fboxsep}{0pt}\colorbox[Hsb]{202, 0.00, 1.0}{\strut cooperating}} {\setlength{\fboxsep}{0pt}\colorbox[Hsb]{202, 0.00, 1.0}{\strut fully}} {\setlength{\fboxsep}{0pt}\colorbox[Hsb]{202, 0.00, 1.0}{\strut with}} {\setlength{\fboxsep}{0pt}\colorbox[Hsb]{202, 0.00, 1.0}{\strut the}} {\setlength{\fboxsep}{0pt}\colorbox[Hsb]{202, 0.00, 1.0}{\strut police}} {\setlength{\fboxsep}{0pt}\colorbox[Hsb]{202, 0.00, 1.0}{\strut .}} {\setlength{\fboxsep}{0pt}\colorbox[Hsb]{202, 0.00, 1.0}{\strut "}}

\textbf{Adversarial}:{\setlength{\fboxsep}{0pt}\colorbox[Hsb]{202, 0.00, 1.0}{\strut @entity1}} {\setlength{\fboxsep}{0pt}\colorbox[Hsb]{202, 0.08, 1.0}{\strut ,}} {\setlength{\fboxsep}{0pt}\colorbox[Hsb]{202, 0.00, 1.0}{\strut @entity2}} {\setlength{\fboxsep}{0pt}\colorbox[Hsb]{202, 0.00, 1.0}{\strut (}} {\setlength{\fboxsep}{0pt}\colorbox[Hsb]{202, 0.00, 1.0}{\strut @entity3}} {\setlength{\fboxsep}{0pt}\colorbox[Hsb]{202, 0.00, 1.0}{\strut )}} {\setlength{\fboxsep}{0pt}\colorbox[Hsb]{202, 0.00, 1.0}{\strut police}} {\setlength{\fboxsep}{0pt}\colorbox[Hsb]{202, 0.00, 1.0}{\strut have}} {\setlength{\fboxsep}{0pt}\colorbox[Hsb]{202, 0.00, 1.0}{\strut arrested}} {\setlength{\fboxsep}{0pt}\colorbox[Hsb]{202, 0.00, 1.0}{\strut four}} {\setlength{\fboxsep}{0pt}\colorbox[Hsb]{202, 0.00, 1.0}{\strut employees}} {\setlength{\fboxsep}{0pt}\colorbox[Hsb]{202, 0.00, 1.0}{\strut of}} {\setlength{\fboxsep}{0pt}\colorbox[Hsb]{202, 0.00, 1.0}{\strut a}} {\setlength{\fboxsep}{0pt}\colorbox[Hsb]{202, 0.00, 1.0}{\strut popular}} {\setlength{\fboxsep}{0pt}\colorbox[Hsb]{202, 0.00, 1.0}{\strut @entity2}} {\setlength{\fboxsep}{0pt}\colorbox[Hsb]{202, 0.00, 1.0}{\strut ethnic}} {\setlength{\fboxsep}{0pt}\colorbox[Hsb]{202, 0.00, 1.0}{\strut -}} {\setlength{\fboxsep}{0pt}\colorbox[Hsb]{202, 0.00, 1.0}{\strut wear}} {\setlength{\fboxsep}{0pt}\colorbox[Hsb]{202, 0.00, 1.0}{\strut chain}} {\setlength{\fboxsep}{0pt}\colorbox[Hsb]{202, 0.00, 1.0}{\strut after}} {\setlength{\fboxsep}{0pt}\colorbox[Hsb]{202, 0.00, 1.0}{\strut a}} {\setlength{\fboxsep}{0pt}\colorbox[Hsb]{202, 0.00, 1.0}{\strut minister}} {\setlength{\fboxsep}{0pt}\colorbox[Hsb]{202, 0.00, 1.0}{\strut spotted}} {\setlength{\fboxsep}{0pt}\colorbox[Hsb]{202, 0.00, 1.0}{\strut a}} {\setlength{\fboxsep}{0pt}\colorbox[Hsb]{202, 0.00, 1.0}{\strut security}} {\setlength{\fboxsep}{0pt}\colorbox[Hsb]{202, 0.00, 1.0}{\strut camera}} {\setlength{\fboxsep}{0pt}\colorbox[Hsb]{202, 0.01, 1.0}{\strut overlooking}} {\setlength{\fboxsep}{0pt}\colorbox[Hsb]{202, 0.00, 1.0}{\strut the}} {\setlength{\fboxsep}{0pt}\colorbox[Hsb]{202, 0.00, 1.0}{\strut changing}} {\setlength{\fboxsep}{0pt}\colorbox[Hsb]{202, 0.00, 1.0}{\strut room}} {\setlength{\fboxsep}{0pt}\colorbox[Hsb]{202, 0.00, 1.0}{\strut of}} {\setlength{\fboxsep}{0pt}\colorbox[Hsb]{202, 0.00, 1.0}{\strut one}} {\setlength{\fboxsep}{0pt}\colorbox[Hsb]{202, 0.00, 1.0}{\strut of}} {\setlength{\fboxsep}{0pt}\colorbox[Hsb]{202, 0.00, 1.0}{\strut its}} {\setlength{\fboxsep}{0pt}\colorbox[Hsb]{202, 0.00, 1.0}{\strut stores}} {\setlength{\fboxsep}{0pt}\colorbox[Hsb]{202, 0.00, 1.0}{\strut .}} {\setlength{\fboxsep}{0pt}\colorbox[Hsb]{202, 0.00, 1.0}{\strut federal}} {\setlength{\fboxsep}{0pt}\colorbox[Hsb]{202, 0.00, 1.0}{\strut education}} {\setlength{\fboxsep}{0pt}\colorbox[Hsb]{202, 0.00, 1.0}{\strut minister}} {\setlength{\fboxsep}{0pt}\colorbox[Hsb]{202, 0.00, 1.0}{\strut @entity13}} {\setlength{\fboxsep}{0pt}\colorbox[Hsb]{202, 0.00, 1.0}{\strut was}} {\setlength{\fboxsep}{0pt}\colorbox[Hsb]{202, 0.00, 1.0}{\strut visiting}} {\setlength{\fboxsep}{0pt}\colorbox[Hsb]{202, 0.00, 1.0}{\strut a}} {\setlength{\fboxsep}{0pt}\colorbox[Hsb]{202, 0.00, 1.0}{\strut @entity15}} {\setlength{\fboxsep}{0pt}\colorbox[Hsb]{202, 0.01, 1.0}{\strut outlet}} {\setlength{\fboxsep}{0pt}\colorbox[Hsb]{202, 0.00, 1.0}{\strut in}} {\setlength{\fboxsep}{0pt}\colorbox[Hsb]{202, 0.00, 1.0}{\strut the}} {\setlength{\fboxsep}{0pt}\colorbox[Hsb]{202, 0.00, 1.0}{\strut tourist}} {\setlength{\fboxsep}{0pt}\colorbox[Hsb]{202, 0.00, 1.0}{\strut resort}} {\setlength{\fboxsep}{0pt}\colorbox[Hsb]{202, 0.00, 1.0}{\strut state}} {\setlength{\fboxsep}{0pt}\colorbox[Hsb]{202, 0.00, 1.0}{\strut of}} {\setlength{\fboxsep}{0pt}\colorbox[Hsb]{202, 0.00, 1.0}{\strut @entity17}} {\setlength{\fboxsep}{0pt}\colorbox[Hsb]{202, 0.00, 1.0}{\strut on}} {\setlength{\fboxsep}{0pt}\colorbox[Hsb]{202, 0.00, 1.0}{\strut friday}} {\setlength{\fboxsep}{0pt}\colorbox[Hsb]{202, 0.00, 1.0}{\strut when}} {\setlength{\fboxsep}{0pt}\colorbox[Hsb]{202, 0.00, 1.0}{\strut she}} {\setlength{\fboxsep}{0pt}\colorbox[Hsb]{202, 0.00, 1.0}{\strut discovered}} {\setlength{\fboxsep}{0pt}\colorbox[Hsb]{202, 0.00, 1.0}{\strut a}} {\setlength{\fboxsep}{0pt}\colorbox[Hsb]{202, 0.00, 1.0}{\strut surveillance}} {\setlength{\fboxsep}{0pt}\colorbox[Hsb]{202, 0.00, 1.0}{\strut camera}} {\setlength{\fboxsep}{0pt}\colorbox[Hsb]{202, 0.00, 1.0}{\strut pointed}} {\setlength{\fboxsep}{0pt}\colorbox[Hsb]{202, 0.00, 1.0}{\strut at}} {\setlength{\fboxsep}{0pt}\colorbox[Hsb]{202, 0.00, 1.0}{\strut the}} {\setlength{\fboxsep}{0pt}\colorbox[Hsb]{202, 0.00, 1.0}{\strut changing}} {\setlength{\fboxsep}{0pt}\colorbox[Hsb]{202, 0.00, 1.0}{\strut room}} {\setlength{\fboxsep}{0pt}\colorbox[Hsb]{202, 0.00, 1.0}{\strut ,}} {\setlength{\fboxsep}{0pt}\colorbox[Hsb]{202, 0.00, 1.0}{\strut police}} {\setlength{\fboxsep}{0pt}\colorbox[Hsb]{202, 0.01, 1.0}{\strut said}} {\setlength{\fboxsep}{0pt}\colorbox[Hsb]{202, 0.00, 1.0}{\strut .}} {\setlength{\fboxsep}{0pt}\colorbox[Hsb]{202, 0.00, 1.0}{\strut four}} {\setlength{\fboxsep}{0pt}\colorbox[Hsb]{202, 0.00, 1.0}{\strut employees}} {\setlength{\fboxsep}{0pt}\colorbox[Hsb]{202, 0.00, 1.0}{\strut of}} {\setlength{\fboxsep}{0pt}\colorbox[Hsb]{202, 0.00, 1.0}{\strut the}} {\setlength{\fboxsep}{0pt}\colorbox[Hsb]{202, 0.00, 1.0}{\strut store}} {\setlength{\fboxsep}{0pt}\colorbox[Hsb]{202, 0.00, 1.0}{\strut have}} {\setlength{\fboxsep}{0pt}\colorbox[Hsb]{202, 0.00, 1.0}{\strut been}} {\setlength{\fboxsep}{0pt}\colorbox[Hsb]{202, 0.00, 1.0}{\strut arrested}} {\setlength{\fboxsep}{0pt}\colorbox[Hsb]{202, 0.00, 1.0}{\strut ,}} {\setlength{\fboxsep}{0pt}\colorbox[Hsb]{202, 0.00, 1.0}{\strut but}} {\setlength{\fboxsep}{0pt}\colorbox[Hsb]{202, 0.00, 1.0}{\strut its}} {\setlength{\fboxsep}{0pt}\colorbox[Hsb]{202, 0.01, 1.0}{\strut manager}} {\setlength{\fboxsep}{0pt}\colorbox[Hsb]{202, 0.00, 1.0}{\strut --}} {\setlength{\fboxsep}{0pt}\colorbox[Hsb]{202, 0.00, 1.0}{\strut herself}} {\setlength{\fboxsep}{0pt}\colorbox[Hsb]{202, 0.00, 1.0}{\strut a}} {\setlength{\fboxsep}{0pt}\colorbox[Hsb]{202, 0.00, 1.0}{\strut woman}} {\setlength{\fboxsep}{0pt}\colorbox[Hsb]{202, 0.00, 1.0}{\strut --}} {\setlength{\fboxsep}{0pt}\colorbox[Hsb]{202, 0.00, 1.0}{\strut was}} {\setlength{\fboxsep}{0pt}\colorbox[Hsb]{202, 0.00, 1.0}{\strut still}} {\setlength{\fboxsep}{0pt}\colorbox[Hsb]{202, 0.00, 1.0}{\strut at}} {\setlength{\fboxsep}{0pt}\colorbox[Hsb]{202, 0.00, 1.0}{\strut large}} {\setlength{\fboxsep}{0pt}\colorbox[Hsb]{202, 0.00, 1.0}{\strut saturday}} {\setlength{\fboxsep}{0pt}\colorbox[Hsb]{202, 0.00, 1.0}{\strut ,}} {\setlength{\fboxsep}{0pt}\colorbox[Hsb]{202, 0.00, 1.0}{\strut said}} {\setlength{\fboxsep}{0pt}\colorbox[Hsb]{202, 0.00, 1.0}{\strut @entity17}} {\setlength{\fboxsep}{0pt}\colorbox[Hsb]{202, 0.00, 1.0}{\strut police}} {\setlength{\fboxsep}{0pt}\colorbox[Hsb]{202, 0.00, 1.0}{\strut superintendent}} {\setlength{\fboxsep}{0pt}\colorbox[Hsb]{202, 0.00, 1.0}{\strut @entity25}} {\setlength{\fboxsep}{0pt}\colorbox[Hsb]{202, 0.00, 1.0}{\strut .}} {\setlength{\fboxsep}{0pt}\colorbox[Hsb]{202, 0.00, 1.0}{\strut state}} {\setlength{\fboxsep}{0pt}\colorbox[Hsb]{202, 0.00, 1.0}{\strut authorities}} {\setlength{\fboxsep}{0pt}\colorbox[Hsb]{202, 0.00, 1.0}{\strut launched}} {\setlength{\fboxsep}{0pt}\colorbox[Hsb]{202, 0.00, 1.0}{\strut their}} {\setlength{\fboxsep}{0pt}\colorbox[Hsb]{202, 0.00, 1.0}{\strut investigation}} {\setlength{\fboxsep}{0pt}\colorbox[Hsb]{202, 0.01, 1.0}{\strut right}} {\setlength{\fboxsep}{0pt}\colorbox[Hsb]{202, 0.00, 1.0}{\strut after}} {\setlength{\fboxsep}{0pt}\colorbox[Hsb]{202, 0.23, 1.0}{\strut @entity13}} {\setlength{\fboxsep}{0pt}\colorbox[Hsb]{202, 0.00, 1.0}{\strut levied}} {\setlength{\fboxsep}{0pt}\colorbox[Hsb]{202, 0.00, 1.0}{\strut her}} {\setlength{\fboxsep}{0pt}\colorbox[Hsb]{202, 0.00, 1.0}{\strut accusation}} {\setlength{\fboxsep}{0pt}\colorbox[Hsb]{202, 0.00, 1.0}{\strut .}} {\setlength{\fboxsep}{0pt}\colorbox[Hsb]{202, 0.00, 1.0}{\strut they}} {\setlength{\fboxsep}{0pt}\colorbox[Hsb]{202, 0.00, 1.0}{\strut found}} {\setlength{\fboxsep}{0pt}\colorbox[Hsb]{202, 0.00, 1.0}{\strut an}} {\setlength{\fboxsep}{0pt}\colorbox[Hsb]{202, 0.00, 1.0}{\strut overhead}} {\setlength{\fboxsep}{0pt}\colorbox[Hsb]{202, 0.00, 1.0}{\strut camera}} {\setlength{\fboxsep}{0pt}\colorbox[Hsb]{202, 0.00, 1.0}{\strut that}} {\setlength{\fboxsep}{0pt}\colorbox[Hsb]{202, 0.00, 1.0}{\strut the}} {\setlength{\fboxsep}{0pt}\colorbox[Hsb]{202, 0.00, 1.0}{\strut minister}} {\setlength{\fboxsep}{0pt}\colorbox[Hsb]{202, 0.00, 1.0}{\strut had}} {\setlength{\fboxsep}{0pt}\colorbox[Hsb]{202, 0.00, 1.0}{\strut spotted}} {\setlength{\fboxsep}{0pt}\colorbox[Hsb]{202, 0.00, 1.0}{\strut and}} {\setlength{\fboxsep}{0pt}\colorbox[Hsb]{202, 0.00, 1.0}{\strut determined}} {\setlength{\fboxsep}{0pt}\colorbox[Hsb]{202, 0.00, 1.0}{\strut that}} {\setlength{\fboxsep}{0pt}\colorbox[Hsb]{202, 0.00, 1.0}{\strut it}} {\setlength{\fboxsep}{0pt}\colorbox[Hsb]{202, 0.00, 1.0}{\strut was}} {\setlength{\fboxsep}{0pt}\colorbox[Hsb]{202, 0.00, 1.0}{\strut indeed}} {\setlength{\fboxsep}{0pt}\colorbox[Hsb]{202, 0.00, 1.0}{\strut able}} {\setlength{\fboxsep}{0pt}\colorbox[Hsb]{202, 0.02, 1.0}{\strut to}} {\setlength{\fboxsep}{0pt}\colorbox[Hsb]{202, 0.00, 1.0}{\strut take}} {\setlength{\fboxsep}{0pt}\colorbox[Hsb]{202, 0.00, 1.0}{\strut photos}} {\setlength{\fboxsep}{0pt}\colorbox[Hsb]{202, 0.00, 1.0}{\strut of}} {\setlength{\fboxsep}{0pt}\colorbox[Hsb]{202, 0.00, 1.0}{\strut customers}} {\setlength{\fboxsep}{0pt}\colorbox[Hsb]{202, 0.00, 1.0}{\strut using}} {\setlength{\fboxsep}{0pt}\colorbox[Hsb]{202, 0.00, 1.0}{\strut the}} {\setlength{\fboxsep}{0pt}\colorbox[Hsb]{202, 0.00, 1.0}{\strut store}} {\setlength{\fboxsep}{0pt}\colorbox[Hsb]{202, 0.00, 1.0}{\strut 's}} {\setlength{\fboxsep}{0pt}\colorbox[Hsb]{202, 0.00, 1.0}{\strut changing}} {\setlength{\fboxsep}{0pt}\colorbox[Hsb]{202, 0.00, 1.0}{\strut room}} {\setlength{\fboxsep}{0pt}\colorbox[Hsb]{202, 0.00, 1.0}{\strut ,}} {\setlength{\fboxsep}{0pt}\colorbox[Hsb]{202, 0.00, 1.0}{\strut according}} {\setlength{\fboxsep}{0pt}\colorbox[Hsb]{202, 0.00, 1.0}{\strut to}} {\setlength{\fboxsep}{0pt}\colorbox[Hsb]{202, 0.00, 1.0}{\strut @entity25}} {\setlength{\fboxsep}{0pt}\colorbox[Hsb]{202, 0.00, 1.0}{\strut .}} {\setlength{\fboxsep}{0pt}\colorbox[Hsb]{202, 0.00, 1.0}{\strut after}} {\setlength{\fboxsep}{0pt}\colorbox[Hsb]{202, 0.00, 1.0}{\strut the}} {\setlength{\fboxsep}{0pt}\colorbox[Hsb]{202, 0.00, 1.0}{\strut incident}} {\setlength{\fboxsep}{0pt}\colorbox[Hsb]{202, 0.00, 1.0}{\strut ,}} {\setlength{\fboxsep}{0pt}\colorbox[Hsb]{202, 0.00, 1.0}{\strut authorities}} {\setlength{\fboxsep}{0pt}\colorbox[Hsb]{202, 0.00, 1.0}{\strut sealed}} {\setlength{\fboxsep}{0pt}\colorbox[Hsb]{202, 0.00, 1.0}{\strut off}} {\setlength{\fboxsep}{0pt}\colorbox[Hsb]{202, 0.00, 1.0}{\strut the}} {\setlength{\fboxsep}{0pt}\colorbox[Hsb]{202, 0.00, 1.0}{\strut store}} {\setlength{\fboxsep}{0pt}\colorbox[Hsb]{202, 0.00, 1.0}{\strut and}} {\setlength{\fboxsep}{0pt}\colorbox[Hsb]{202, 0.00, 1.0}{\strut summoned}} {\setlength{\fboxsep}{0pt}\colorbox[Hsb]{202, 0.00, 1.0}{\strut six}} {\setlength{\fboxsep}{0pt}\colorbox[Hsb]{202, 0.00, 1.0}{\strut top}} {\setlength{\fboxsep}{0pt}\colorbox[Hsb]{202, 0.00, 1.0}{\strut officials}} {\setlength{\fboxsep}{0pt}\colorbox[Hsb]{202, 0.00, 1.0}{\strut from}} {\setlength{\fboxsep}{0pt}\colorbox[Hsb]{202, 0.00, 1.0}{\strut @entity15}} {\setlength{\fboxsep}{0pt}\colorbox[Hsb]{202, 0.00, 1.0}{\strut ,}} {\setlength{\fboxsep}{0pt}\colorbox[Hsb]{202, 0.02, 1.0}{\strut he}} {\setlength{\fboxsep}{0pt}\colorbox[Hsb]{202, 0.00, 1.0}{\strut said}} {\setlength{\fboxsep}{0pt}\colorbox[Hsb]{202, 0.00, 1.0}{\strut .}} {\setlength{\fboxsep}{0pt}\colorbox[Hsb]{202, 0.00, 1.0}{\strut the}} {\setlength{\fboxsep}{0pt}\colorbox[Hsb]{202, 0.00, 1.0}{\strut arrested}} {\setlength{\fboxsep}{0pt}\colorbox[Hsb]{202, 0.00, 1.0}{\strut staff}} {\setlength{\fboxsep}{0pt}\colorbox[Hsb]{202, 0.00, 1.0}{\strut have}} {\setlength{\fboxsep}{0pt}\colorbox[Hsb]{202, 0.00, 1.0}{\strut been}} {\setlength{\fboxsep}{0pt}\colorbox[Hsb]{202, 0.00, 1.0}{\strut charged}} {\setlength{\fboxsep}{0pt}\colorbox[Hsb]{202, 0.00, 1.0}{\strut with}} {\setlength{\fboxsep}{0pt}\colorbox[Hsb]{202, 0.00, 1.0}{\strut voyeurism}} {\setlength{\fboxsep}{0pt}\colorbox[Hsb]{202, 0.00, 1.0}{\strut and}} {\setlength{\fboxsep}{0pt}\colorbox[Hsb]{202, 0.00, 1.0}{\strut breach}} {\setlength{\fboxsep}{0pt}\colorbox[Hsb]{202, 0.00, 1.0}{\strut of}} {\setlength{\fboxsep}{0pt}\colorbox[Hsb]{202, 0.00, 1.0}{\strut privacy}} {\setlength{\fboxsep}{0pt}\colorbox[Hsb]{202, 0.00, 1.0}{\strut ,}} {\setlength{\fboxsep}{0pt}\colorbox[Hsb]{202, 0.00, 1.0}{\strut according}} {\setlength{\fboxsep}{0pt}\colorbox[Hsb]{202, 0.00, 1.0}{\strut to}} {\setlength{\fboxsep}{0pt}\colorbox[Hsb]{202, 0.00, 1.0}{\strut the}} {\setlength{\fboxsep}{0pt}\colorbox[Hsb]{202, 0.00, 1.0}{\strut police}} {\setlength{\fboxsep}{0pt}\colorbox[Hsb]{202, 0.00, 1.0}{\strut .}} {\setlength{\fboxsep}{0pt}\colorbox[Hsb]{202, 0.00, 1.0}{\strut if}} {\setlength{\fboxsep}{0pt}\colorbox[Hsb]{202, 0.00, 1.0}{\strut convicted}} {\setlength{\fboxsep}{0pt}\colorbox[Hsb]{202, 0.00, 1.0}{\strut ,}} {\setlength{\fboxsep}{0pt}\colorbox[Hsb]{202, 0.00, 1.0}{\strut they}} {\setlength{\fboxsep}{0pt}\colorbox[Hsb]{202, 0.00, 1.0}{\strut could}} {\setlength{\fboxsep}{0pt}\colorbox[Hsb]{202, 0.00, 1.0}{\strut spend}} {\setlength{\fboxsep}{0pt}\colorbox[Hsb]{202, 0.00, 1.0}{\strut up}} {\setlength{\fboxsep}{0pt}\colorbox[Hsb]{202, 0.00, 1.0}{\strut to}} {\setlength{\fboxsep}{0pt}\colorbox[Hsb]{202, 0.00, 1.0}{\strut three}} {\setlength{\fboxsep}{0pt}\colorbox[Hsb]{202, 0.00, 1.0}{\strut years}} {\setlength{\fboxsep}{0pt}\colorbox[Hsb]{202, 0.00, 1.0}{\strut in}} {\setlength{\fboxsep}{0pt}\colorbox[Hsb]{202, 0.00, 1.0}{\strut jail}} {\setlength{\fboxsep}{0pt}\colorbox[Hsb]{202, 0.00, 1.0}{\strut ,}} {\setlength{\fboxsep}{0pt}\colorbox[Hsb]{202, 0.00, 1.0}{\strut @entity25}} {\setlength{\fboxsep}{0pt}\colorbox[Hsb]{202, 0.00, 1.0}{\strut said}} {\setlength{\fboxsep}{0pt}\colorbox[Hsb]{202, 0.01, 1.0}{\strut .}} {\setlength{\fboxsep}{0pt}\colorbox[Hsb]{202, 0.00, 1.0}{\strut officials}} {\setlength{\fboxsep}{0pt}\colorbox[Hsb]{202, 0.00, 1.0}{\strut from}} {\setlength{\fboxsep}{0pt}\colorbox[Hsb]{202, 0.00, 1.0}{\strut @entity15}} {\setlength{\fboxsep}{0pt}\colorbox[Hsb]{202, 0.00, 1.0}{\strut --}} {\setlength{\fboxsep}{0pt}\colorbox[Hsb]{202, 0.00, 1.0}{\strut which}} {\setlength{\fboxsep}{0pt}\colorbox[Hsb]{202, 0.00, 1.0}{\strut sells}} {\setlength{\fboxsep}{0pt}\colorbox[Hsb]{202, 0.00, 1.0}{\strut ethnic}} {\setlength{\fboxsep}{0pt}\colorbox[Hsb]{202, 0.00, 1.0}{\strut garments}} {\setlength{\fboxsep}{0pt}\colorbox[Hsb]{202, 0.00, 1.0}{\strut ,}} {\setlength{\fboxsep}{0pt}\colorbox[Hsb]{202, 0.00, 1.0}{\strut fabrics}} {\setlength{\fboxsep}{0pt}\colorbox[Hsb]{202, 0.00, 1.0}{\strut and}} {\setlength{\fboxsep}{0pt}\colorbox[Hsb]{202, 0.00, 1.0}{\strut other}} {\setlength{\fboxsep}{0pt}\colorbox[Hsb]{202, 0.00, 1.0}{\strut products}} {\setlength{\fboxsep}{0pt}\colorbox[Hsb]{202, 0.00, 1.0}{\strut --}} {\setlength{\fboxsep}{0pt}\colorbox[Hsb]{202, 0.00, 1.0}{\strut are}} {\setlength{\fboxsep}{0pt}\colorbox[Hsb]{202, 0.00, 1.0}{\strut heading}} {\setlength{\fboxsep}{0pt}\colorbox[Hsb]{202, 0.00, 1.0}{\strut to}} {\setlength{\fboxsep}{0pt}\colorbox[Hsb]{202, 0.00, 1.0}{\strut @entity17}} {\setlength{\fboxsep}{0pt}\colorbox[Hsb]{202, 0.00, 1.0}{\strut to}} {\setlength{\fboxsep}{0pt}\colorbox[Hsb]{202, 0.00, 1.0}{\strut work}} {\setlength{\fboxsep}{0pt}\colorbox[Hsb]{202, 0.00, 1.0}{\strut with}} {\setlength{\fboxsep}{0pt}\colorbox[Hsb]{202, 0.00, 1.0}{\strut investigators}} {\setlength{\fboxsep}{0pt}\colorbox[Hsb]{202, 0.00, 1.0}{\strut ,}} {\setlength{\fboxsep}{0pt}\colorbox[Hsb]{202, 0.00, 1.0}{\strut according}} {\setlength{\fboxsep}{0pt}\colorbox[Hsb]{202, 0.00, 1.0}{\strut to}} {\setlength{\fboxsep}{0pt}\colorbox[Hsb]{202, 0.00, 1.0}{\strut the}} {\setlength{\fboxsep}{0pt}\colorbox[Hsb]{202, 0.00, 1.0}{\strut company}} {\setlength{\fboxsep}{0pt}\colorbox[Hsb]{202, 0.00, 1.0}{\strut .}} {\setlength{\fboxsep}{0pt}\colorbox[Hsb]{202, 0.00, 1.0}{\strut "}} {\setlength{\fboxsep}{0pt}\colorbox[Hsb]{202, 0.00, 1.0}{\strut @entity15}} {\setlength{\fboxsep}{0pt}\colorbox[Hsb]{202, 0.00, 1.0}{\strut is}} {\setlength{\fboxsep}{0pt}\colorbox[Hsb]{202, 0.01, 1.0}{\strut deeply}} {\setlength{\fboxsep}{0pt}\colorbox[Hsb]{202, 0.00, 1.0}{\strut concerned}} {\setlength{\fboxsep}{0pt}\colorbox[Hsb]{202, 0.00, 1.0}{\strut and}} {\setlength{\fboxsep}{0pt}\colorbox[Hsb]{202, 0.00, 1.0}{\strut shocked}} {\setlength{\fboxsep}{0pt}\colorbox[Hsb]{202, 0.00, 1.0}{\strut at}} {\setlength{\fboxsep}{0pt}\colorbox[Hsb]{202, 0.00, 1.0}{\strut this}} {\setlength{\fboxsep}{0pt}\colorbox[Hsb]{202, 0.00, 1.0}{\strut allegation}} {\setlength{\fboxsep}{0pt}\colorbox[Hsb]{202, 0.00, 1.0}{\strut ,}} {\setlength{\fboxsep}{0pt}\colorbox[Hsb]{202, 0.00, 1.0}{\strut "}} {\setlength{\fboxsep}{0pt}\colorbox[Hsb]{202, 0.00, 1.0}{\strut the}} {\setlength{\fboxsep}{0pt}\colorbox[Hsb]{202, 0.01, 1.0}{\strut company}} {\setlength{\fboxsep}{0pt}\colorbox[Hsb]{202, 0.00, 1.0}{\strut said}} {\setlength{\fboxsep}{0pt}\colorbox[Hsb]{202, 0.00, 1.0}{\strut in}} {\setlength{\fboxsep}{0pt}\colorbox[Hsb]{202, 0.00, 1.0}{\strut a}} {\setlength{\fboxsep}{0pt}\colorbox[Hsb]{202, 0.00, 1.0}{\strut statement}} {\setlength{\fboxsep}{0pt}\colorbox[Hsb]{202, 0.00, 1.0}{\strut .}} {\setlength{\fboxsep}{0pt}\colorbox[Hsb]{202, 0.00, 1.0}{\strut "}} {\setlength{\fboxsep}{0pt}\colorbox[Hsb]{202, 0.00, 1.0}{\strut we}} {\setlength{\fboxsep}{0pt}\colorbox[Hsb]{202, 0.00, 1.0}{\strut are}} {\setlength{\fboxsep}{0pt}\colorbox[Hsb]{202, 0.00, 1.0}{\strut in}} {\setlength{\fboxsep}{0pt}\colorbox[Hsb]{202, 0.00, 1.0}{\strut the}} {\setlength{\fboxsep}{0pt}\colorbox[Hsb]{202, 0.00, 1.0}{\strut process}} {\setlength{\fboxsep}{0pt}\colorbox[Hsb]{202, 0.00, 1.0}{\strut of}} {\setlength{\fboxsep}{0pt}\colorbox[Hsb]{202, 0.00, 1.0}{\strut investigating}} {\setlength{\fboxsep}{0pt}\colorbox[Hsb]{202, 0.00, 1.0}{\strut this}} {\setlength{\fboxsep}{0pt}\colorbox[Hsb]{202, 0.00, 1.0}{\strut internally}} {\setlength{\fboxsep}{0pt}\colorbox[Hsb]{202, 0.00, 1.0}{\strut and}} {\setlength{\fboxsep}{0pt}\colorbox[Hsb]{202, 0.00, 1.0}{\strut will}} {\setlength{\fboxsep}{0pt}\colorbox[Hsb]{202, 0.00, 1.0}{\strut be}} {\setlength{\fboxsep}{0pt}\colorbox[Hsb]{202, 0.00, 1.0}{\strut cooperating}} {\setlength{\fboxsep}{0pt}\colorbox[Hsb]{202, 0.00, 1.0}{\strut fully}} {\setlength{\fboxsep}{0pt}\colorbox[Hsb]{202, 0.01, 1.0}{\strut with}} {\setlength{\fboxsep}{0pt}\colorbox[Hsb]{202, 0.00, 1.0}{\strut the}} {\setlength{\fboxsep}{0pt}\colorbox[Hsb]{202, 0.00, 1.0}{\strut police}} {\setlength{\fboxsep}{0pt}\colorbox[Hsb]{202, 0.00, 1.0}{\strut .}} {\setlength{\fboxsep}{0pt}\colorbox[Hsb]{202, 0.00, 1.0}{\strut "}}
\textbf{$\Delta\hat{y}$}: \emph{0.005}


\end{document}